\documentclass[Afour,sageh,times]{sagej}

\usepackage{moreverb,url}

\usepackage{siunitx}
\usepackage{dsfont}
\usepackage{algorithm}
\usepackage{algpseudocode}
\usepackage{subcaption}
\usepackage[colorlinks,bookmarksopen,bookmarksnumbered,citecolor=red,urlcolor=red]{hyperref}

\newcommand\BibTeX{{\rmfamily B\kern-.05em \textsc{i\kern-.025em b}\kern-.08em
T\kern-.1667em\lower.7ex\hbox{E}\kern-.125emX}}

\begin{document}

% \runninghead{Rudin}

% \title{Agile Locomotion in the Wild Using Multi-expert Distillation and RL Fine-tuning}
% \title{Parkour in the Wild: Learning a General Agile Locomotion Policy Using Multi-expert Distillation and RL Fine-tuning}
\title{Parkour in the wild: Learning a general and extensible agile locomotion policy using multi-expert distillation and RL Fine-tuning}
% \itshape{SAGE Publications}}

\author{Nikita Rudin\affilnum{1,2}, Junzhe He\affilnum{1}, Joshua Aurand\affilnum{1}, and Marco Hutter\affilnum{1}}

\affiliation{\affilnum{1}Robotic Systems Lab, ETH Zurich\\
\affilnum{2}NVIDIA Switzerland}

\corrauth{Nikita Rudin}

\email{rudinn@ethz.ch}

\begin{abstract}
Legged robots are well-suited for navigating terrains inaccessible to wheeled robots, making them ideal for applications in search and rescue or space exploration. However, current control methods often struggle to generalize across diverse, unstructured environments. This paper introduces a novel framework for agile locomotion of legged robots by combining multi-expert distillation with reinforcement learning (RL) fine-tuning to achieve robust generalization.
Initially, terrain-specific expert policies are trained to develop specialized locomotion skills. These policies are then distilled into a unified foundation policy via the DAgger algorithm. The distilled policy is subsequently fine-tuned using RL on a broader terrain set, including real-world 3D scans. The framework allows further adaptation to new terrains through repeated fine-tuning.
The proposed policy leverages depth images as exteroceptive inputs, enabling robust navigation across diverse, unstructured terrains. Experimental results demonstrate significant performance improvements over existing methods in synthesizing multi-terrain skills into a single controller. Deployment on the ANYmal D robot validates the policy's ability to navigate complex environments with agility and robustness, setting a new benchmark for legged robot locomotion.
\end{abstract}

% \begin{abstract}
% Legged robots have the potential to navigate terrains inaccessible to wheeled robots, making them ideal for applications like search and rescue and space exploration. However, current control methods are limited in their ability to generalize across diverse and unstructured environments. This paper presents a novel approach to developing a general agile locomotion policy for legged robots using a combination of multi-expert distillation and reinforcement learning (RL) fine-tuning. 
% Initially, expert policies are trained separately on distinct terrains, each requiring specialized locomotion skills. These experts are then distilled into a single policy using the DAgger algorithm. Finally, the distilled policy is used as the foundation for fine-tuning with RL on a broader set of terrains, including real-world 3D scans. The foundation policy can be further adapted to new terrains with repeated fine-tuning.
% The resulting policy directly uses depth images as exteroceptive observation, enabling robust navigation across various unstructured terrains. Our approach outperforms existing methods in combining multiple skills into a single controller, demonstrating significant improvements in both seen and unseen terrains. The policy is successfully deployed on the ANYmal D robot, showcasing its capability to handle complex environments with agility and robustness. 
% \end{abstract}

\keywords{Legged Robots, Reinforcement Learning, Distillation}

\maketitle
% List of figures
% 1: big figure with real-world deployment
% 2: method illustration
% 3: expert skills/terrains + modification for distillation
% 4: noise model depth cameras
% 5: field of view of cameras vs elevation map
% 6: climbing field of view
% 7: policy architecture (maybe included in method illustration?)

% List of tables:
% 1: performance results
\begin{figure*}[ht]
\includegraphics[width=\linewidth]{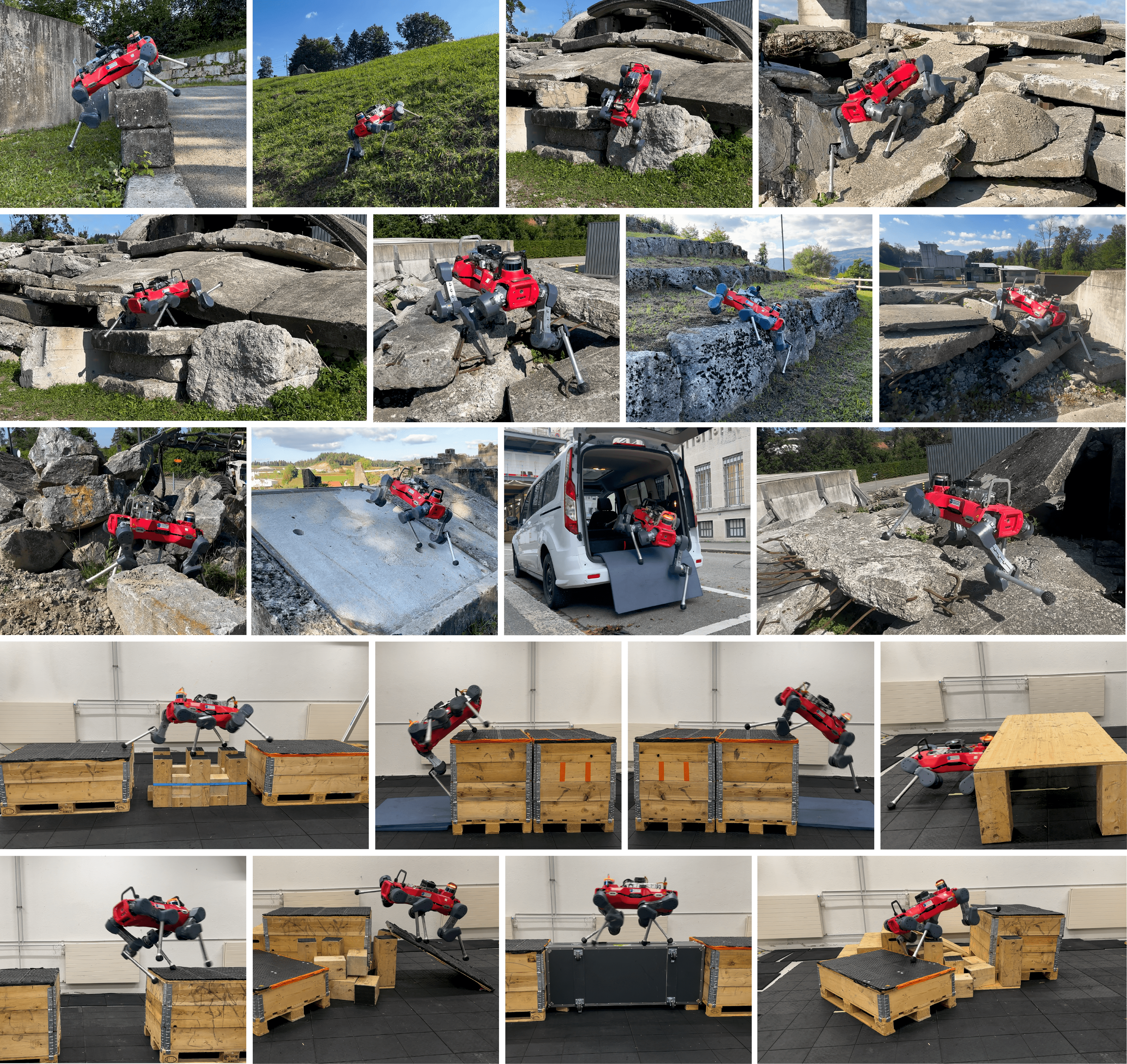}
\label{fig:cover}
\caption{Deployment of our policy on the ANYmal D robot in real-world environments. The robot performs a variety of motions to cross the different unstructured obstacles. A single policy is used in all deployments. }
\end{figure*}
\section{Introduction}
\label{sec:introduction}
Legged robots have long been promising to conquer terrains inaccessible to their wheeled counterparts. While we saw tremendous progress in the field, these robots are still far from matching the performance of humans or animals in complex environments. Due to this gap, applications such as search and rescue and space exploration, which would benefit the most from a versatile legged robot remain impractical.  While robotic hardware continues to evolve, software is the current bottleneck and is far from using the hardware to its full potential.

Control methods have been developing quickly in recent years. The democratization of reinforcement learning (RL) has allowed robots to get out of the lab and tackle unstructured terrains with unprecedented robustness \cite{Lee_2020, Miki_2022, choi_23}. More recently, researchers have been using RL to train agile policies capable of traversing complex obstacles \cite{extremeparkour, zhuang2023robotparkourlearning, hoeller2023anymal}. 

These works demonstrated the maturity of RL and its capacity to solve various specific tasks. Still, the resulting policies can only be applied in the narrow domain where they were trained. For each new task, a policy is typically trained from scratch without any benefits from the knowledge contained in previously trained policies. We lack a system capable of re-using and combining the abilities of individual skills into a general controller, which would preserve the performance of individual skills but also be adaptable to new scenarios. The system must be easily extensible to new tasks. As the robot is deployed in the real world,  there will be unforeseen scenarios that the controller will not be able to handle successfully. The pipeline must facilitate training the controller for the missing skills while preventing it from forgetting any of the previously acquired knowledge.

In the context of legged locomotion, skills represent motions required to cross different terrains and obstacles. A general controller would then be capable of crossing a large variety of terrains using a diverse set of motions and adapting to previously unseen complex and unstructured scenarios.

% Various ways have been explored to combine policies, from heuristic-based switching (cite ??) to distillation (barkour, stanford parkour) and hierarchical RL (anymal parkkour + ?). Unfortunately, none of the proposed methods have demonstrated a scalable solution capable of combining a large number of underlying skills and adapting to previously unseen scenarios.

% In this work, we revisit the problem of skill combination for agile locomotion on unstructured terrain. In particular, we are interested in training fundamental skills and then developing a method that can use those expert skills and generalize to new unseen tasks without requiring training an expert for each possible scenario. In our case, tasks correspond to terrains requiring different motions and generalization corresponds to reusing those motions in more complex unstructured terrains. 

As an additional challenge, policies trained for agile locomotion must perceive the obstacles they are navigating. Standard methods of terrain reconstruction are insufficient since they rely on precise state estimation, which is unavailable during agile motions. Furthermore, parts of the obstacles often remain outside the sensors' field of view during the approach phase. Reconstructing the terrain is not possible in that case, but the policy must nevertheless execute the correct motion.
% Providing elevation data of the terrain in the vicinity of the robot is a standard way to provide sufficient information for the training of single-skill policies (cite RSL, CMU, Stanford, MIT). However, this approach has two main limitations. First, the combined controller must be able to distinguish different obstacles which is not possible using a small-scale top-down view. Furthermore, reconstructing the top-down view requires information that might be outside of the field of view of the sensors. To circumvent these problems, we examine the feasibility of directly using depth camera images as inputs to the locomotion policy.
\subsection{Contribution}  
We propose a three-stage approach to provide quadrupedal robots with unprecedented locomotion capabilities. First, the different experts are trained separately using RL. 

We then examine different methods to combine them into a common controller. We compare our previously developed hierarchical approach (\cite{hoeller2023anymal}), skill encoding through a variational auto-encoder, and multi-expert distillation. We find that in their standard form, none of the methods achieve reasonable performance on complex terrains for which expert policies are unavailable. Hierarchical and latent-encoding approaches tend to fall into local minima solving only subsets of terrains and failing to generalize to more complex obstacles. Distillation achieves non-trivial performance on all terrains, but due to the multi-modal nature of the problem, its performance is inferior to that of individual experts, and its generalization to unseen terrains is limited. However, a distilled policy can be considered as a foundation model and repeatedly fine-tuned with RL. This training is done on various terrains, combining the ones used during distillation with new ones that were not seen during the training of experts. 

We show that after fine-tuning on a diverse set of terrains, the policy’s performance increases on basic terrains, and it learns to solve more complex terrains for which experts could not be trained directly. Interestingly, it even learns new behaviors, for example, to improve the visibility of obstacles in the depth camera's field of view by adapting the motion of the body. A policy trained in this way can be successfully deployed on new terrains that were not seen during training.

Furthermore, we show that new skills can be added using repeated fine-tuning by simply adding new terrains to the training set. The policies learn unique motions to solve new obstacles while maintaining their performance on previously seen tasks. 
% They are then distilled into a single policy using DAgger (~\cite{dagger}), where we iteratively train a policy on a dataset of observation-action pairs from expert demonstrations using supervised learning. The distilled policy is capable of crossing different obstacles by imitating the motions of the experts. However, due to the multi-modal nature of the problem, its performance is inferior to the one of individual experts, and its generalization to unseen terrains is limited. To recover the highest performance and achieve a general policy, we then fine-tune the distilled policy using RL. This training is done on various terrains, combining the ones used during distillation with new ones that were not seen during the training of experts.

To solve the perception problem, we examine the feasibility of directly using depth camera images as inputs to the locomotion policy. We show that, with a carefully designed noise model, this approach yields a robust perceptive locomotion policy that uses only depth images and allows the robot to cross a variety of unstructured terrains with large obstacles and navigate complex environments in ways that were previously considered unachievable.

We test the resulting policy extensively in both indoor and outdoor scenarios. We deploy the robot in search and rescue training facilities, where we test the generalization of the policy across unseen piles of rubble, rocks, and collapsed buildings. These deployments also demonstrate the robustness of the policy to degraded perception (tall grass, changing luminosity, reflections, direct sunlight), and physical disturbances such as slippery terrain (gravel, mud), unstable obstacles (rolling rocks, unstable rubble), and foot traps (concrete rebar, small cracks).

With this, we demonstrate the first scalable training approach for legged locomotion that is able to handle increasing complexity by repeated fine-tuning without a significant loss in performance. This training approach results in a complete locomotion controller with an unparalleled level of agility, robustness, and generalization to unseen scenarios.

\subsection{Related work}
\label{subsec:related-work}
% Locomotion on complex terrains: (RL (wild animal, advanced skills end-to-end, Chong risky terrains, something from CMU, Stanford)+ MPC (Ruben Grandia’s T-RO paper, MIT mini-cheetah) + RL&MPC (Deep tracking control) 
\subsubsection{Legged locomotion}:
Model-Predictive Control (MPC) has been used extensively to control legged robots on different terrains \cite{grandia2022perceptivelocomotionnonlinearmodel, jenelten2022Tamols, kim2020loco}. In particular, it can successfully solve high-precision tasks such as careful foot placement on sparse or narrow obstacles. MPC, however, relies on high-quality models and can fail due to noisy terrain reconstruction or slippage. Additionally, MPC makes use of highly constrained contact schedules which are suitable for walking but can inhibit more complex types of motion. RL presents a promising alternative to the more traditional model-based approaches, requiring fewer assumptions and heuristics. Typically relying on simulation, RL is used to explore and find control policies through a trial-and-error process. While such learned policies can achieve good performance even on challenging terrains \cite{rudin2022advanced}, \cite{zhang2024learningagilelocomotionrisky}, \cite{extremeparkour}, \cite{tencent} there is still a considerable performance gap compared to MPC in high-precision tasks and extensive work is required to train each motion. Finally, researchers have explored hybrid approaches combining MPC with RL in different ways to bypass the limitations of each method. ~\cite{rloc, glide} use DRL to predict footholds that are then tracked by model-based controllers. \cite{deeptracking, kang_RL+MPC} utilize model-based approaches to generate optimal trajectories that are then tracked or imitated by an RL policy. Unfortunately, while such hybrid approaches can combine the strengths of both methods, they also tend to combine the weaknesses. In practice, these approaches retain the limiting assumptions of MPC while also requiring the extensive tuning effort of RL.

% perception
While robust walking can rely purely on proprioception \cite{Lee_2020}, perception of the environment is needed to overcome more challenging obstacles. Elevation maps are a standard method to provide a robot with information about its surrounding terrain \cite{frankhauserElevationMap, miki2022elevationmappinglocomotionnavigation}. This approach is fundamentally limited to 2D and cannot represent over-hanging obstacles. Additionally, it is sensitive to state-estimation drift results on a map too imprecise to be used for crossing challenging obstacles. Hence, \cite{hoeller2023anymal} proposes a neural terrain reconstruction to achieve accurate 3D terrain representations. The approach can accurately reconstruct obstacles under state estimation drift, noisy measurements, and partial observability. Unfortunately, the performance drops significantly in unstructured terrains since it can not properly guess unseen parts of obstacles. Instead, recent works have shown that a locomotion policy can learn to use the output of depth cameras. The training of the policy can be done directly with end-to-end RL \cite{yang2022learning, yu_visual_locomotion}, but most works (\cite{agarwal2022legged, extremeparkour, zhuang2023robotparkourlearning}) choose to use a teacher-student approach, where a teacher is first trained with privileged information, then a student policy is trained to imitate the actions of the teacher using supervised training. The student policy does not have access to the privileged information and must reconstruct it using depth images with noise models and further processing to bridge the sim-to-real gap. 

% Combination of skills: Stanford parkour, Barkour, ANYmal parkour, something from manipulation (??), imitation + diffusion (TRI/Tedrake??)

Previously described approaches can produce agile and robust motions, but additional mechanisms are required to combine these motions into a common control strategy. Recently, the \textit{Parkour} setting has proven to be a great benchmark for the combination of multiple agile motions. In that setting, the robot must cross multiple obstacles, each requiring a different motion or skill. The robot does not know the arrangements of obstacles in advance and must, therefore, adapt its behavior based on its perceptive inputs. Various approaches have tackled this challenge. \cite{hoeller2023anymal} presents a hierarchical formulation where a high-level policy selects lower-level expert skills based on environment perception. In contrast to that, \cite{extremeparkour} trains a single policy on different obstacles, while \cite{caluwaerts2023barkour} and \cite{zhuang2023robotparkourlearning} use a distillation of expert policies to train a single policy capable of performing all skills, similar to the first step of our proposed approach. Those works, however, only train a limited number of new skills and do not show signs of generalization to new terrains.
% Foundation models and generalization in robotics: RT-X, Levine’s lab

In the broader machine learning community, foundation models - models trained on a large variety of data capable of performing many different tasks (\cite{GPT}, \cite{alayrac2022flamingovisuallanguagemodel}), have shown to be able to generalize to untrained scenarios with ease. However, such models are not yet as common in robotics. PaLM-E (\cite{driess2023palme}) is a language model that is trained on multi-modal inputs, including robotics data. It is capable of reasoning about different modalities and acting as a high-level control policy. \cite{PAFF} proposed to fine-tune policies in new environments by using a pre-trained foundation model to label the demonstrations. Utilizing a large dataset collected in 17 months on 13 robots, Robotics Transformer (RT-1 \cite{rt1}) takes multi-modal inputs to generate joint actions that are shown to be capable of robustly generalizing to untrained tasks. A follow-up work, Robotic Transformer 2 (RT2 \cite{rt2}), demonstrates a generalized robot control framework by fine-tuning a pre-trained vision-language model on robot trajectories, resulting in a vision-language-action (VLA) model that enables more complicated semantic reasoning. Building on RT-2, RT-X (\cite{rtx}), introduces X-embodiment capabilities -- it learns from a diverse array of robot types and environments across various datasets. This cross-platform learning allows RT-X to generalize skills across different robot platforms (e.g., robotic arms, drones) and adapt better to new contexts, effectively bridging the embodiment gap and enhancing robustness in real-world applications. \cite{gupta2022metamorph} proposes a foundation model capable of controlling different robots across a fixed design space.

\section{Method} 
\label{sec:method}
\begin{figure*}[ht]
\includegraphics[width=\linewidth]{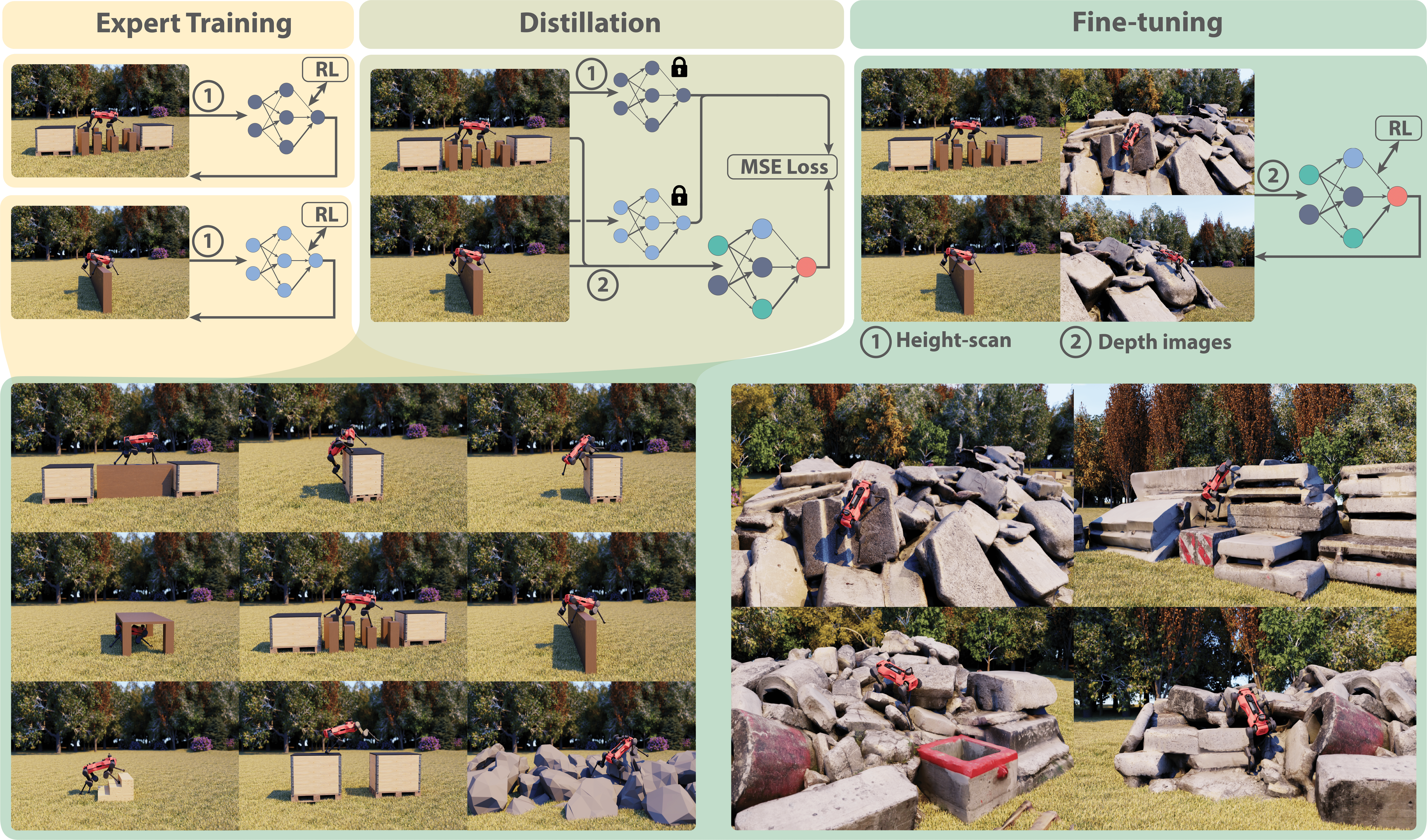}
\label{fig:training-pipeline}
\caption{Description of our approach. The training is decomposed into three stages. First, 9 individual skills are trained separately using RL. The skills use elevation maps as perceptive input. Second, the skills are distilled into a single network using supervised learning and the Dagger approach. The distilled policy receives depth images instead of elevation maps. We use the 9 skills and terrains of the first phase during distillation. Finally, the distilled policy is fine-tuned using RL. During this phase, there is no supervision from experts, and the policy is trained on a combination of the 9 terrains of the previous phase as well as 15 new terrains, which are obtained from 3D scans of real-world piles of rubble in search-and-rescue training facilities. Only 4 out of 15 scans are shown in the figure.}
\end{figure*}

\begin{table}[t]
\centering
\caption{Symbols.}
\label{tab:symbols}
\begin{tabular}{p{0.15\linewidth} p{0.7\linewidth}}
\hline
\textbf{Symbol}      & \textbf{Description}       \\ \hline
$\mathbf{r}$, $\mathbf{r}^*$ & Current and target base positions \\
$\psi$, $\psi^*$ & Current and target base headings \\
$t^*$  & Remaining time to reach the target \\
$\alpha$ & Angle between base z-axis and gravity \\
$\mathbf{v}_b$, $\boldsymbol{\omega}_b$ & Base velocities in base frame \\
$\mathbf{g}_b$ & Gravity vector in base frame \\
$\mathbf{q}, \dot{\mathbf{q}}, \boldsymbol{\tau}$ & Joint positions and  velocities\\ 
$\mathbf{q}_{\lim}, \dot{\mathbf{q}}_{\lim}$ & Joint limits\\ 
$\mathbf{q}^*, \mathbf{q}_d$ & Desired and default joint positions\\
$\boldsymbol{\tau}$, $\boldsymbol{\tau}_{\lim}$ & Joint Torques and torque limits \\
$\mathbf{v}_f$, $\mathbf{F}_f$ & Feet linear velocity and contact force \\
$\mathbf{Em}$ & Elevation map around the robot \\
$\mathbf{Ls}$ & Lidar (horizontal) scan around the robot \\
$\mathbf{I}$ & Depth images  \\
$\mathds{S}_L$ & Target reached,  \\
{} & $ \mathds{S}_L=\mathds{1}_{\lVert \mathbf{r}_{xy} - \mathbf{r}^*_{xy}\rVert < 0.25} \mathds{1}_{\lVert \psi - \psi^* \rVert < 0.5}$\\ 
\hline\\
\end{tabular}
\end{table}

\begin{table}[t]
\def\num#1{\numx#1}\def\numx#1e#2{{#1}e{#2}}
\centering
\caption{Rewards used during fine-tuning.}
\label{tab:reward}
\begin{tabular}{p{0.25\linewidth} p{0.5\linewidth} p{0.1\linewidth}}
\hline
\textbf{Reward Term} & \textbf{Expression} & \textbf{Weight}\\
\hline
Track position &$ \mathds{1}_{t^*<1}(1 - 0.5\lVert\mathbf{r}_{xy}-\mathbf{r}_{xy}^*\rVert)$  & 10\\ 
Track heading & $ \mathds{1}_{t^*<1}(1 - 0.5\lVert\psi-\psi^*\rVert)$  & 5\\ 
Joint velocity & $\lVert \dot{\mathbf{q}} \rVert ^2$ & \num{-1e-3} \\ 
Torque & $\lVert \boldsymbol{\tau}\rVert^2$ & \num{-1e-5}\\ 
Joint vel. limit & $\sum_{i=1}^{12} \max (\lvert\dot{\mathbf{q}}_{i}\rvert-\dot{q}_{\lim},0)$ & -1  \\ 
Torque limit & $ \sum_{i=1}^{12} \max (\lvert\boldsymbol{\tau}_{i}\rvert-\tau_{\lim},0)$ & -0.2 \\ 
Base acc. &$\lVert \dot{\mathbf{v}} \rVert ^2 + 0.02\lVert \dot{\boldsymbol{\omega}} \rVert ^2$ & \num{-1e-3} \\ 
Feet acc.   &$ \sum_{f=1}^4 \lVert \dot{\mathbf{v}}_{f}\rVert$ & \num{-2e-3} \\ 
Action rate   &$\lVert \mathbf{q}^*_t-\mathbf{q}^*_{t-1} \Vert ^2$ & \num{-1e-2} \\ 
Feet force  & $\sum_{f=1}^4 \max(\lVert F_{f} \rVert - 700, 0)^2$ & \num{-1e-5} \\ 
Don't wait & $ \mathds{1}(\lVert \mathbf{v}_b \rVert < 0.2)$ & -1 \\ 
Stand at target & $\mathds{S}_L \lVert \mathbf{q} - \mathbf{q}_d\rVert$ & -0.5\\ 
Collision & $\mathds{1}_{\textit{knee/shank collision}}$ & -1 \\ 
Termination & $\mathds{1}_{\alpha>\SI{135}{\degree}} + \mathds{1}_{\dot{\mathbf{q}}>\dot{\mathbf{q}}_{\lim}}$ & \num{-2e3}\\
\hline
\end{tabular}
\end{table}

\begin{table}[t]
\centering
\caption{Observations.}
\label{tab:observations}
\begin{tabular}{p{0.25\linewidth} p{0.15\linewidth} p{0.15\linewidth} p{0.15\linewidth}}
\hline
 \textbf{Observation} & \textbf{Expert}  & \textbf{Student}  & \textbf{Critic}    \\ \hline
% \textbf{Common}      & \\ 
$\mathbf{v}_b$ & $\times$ & $\times$ & $\times$\\
$\boldsymbol{\omega}_b$ & $\times$ & $\times$ & $\times$\\
$\mathbf{g}_b$ & $\times$ & $\times$ & $\times$\\
$\mathbf{q}, \dot{\mathbf{q}}$ & $\times$ & $\times$ & $\times$\\
$\mathbf{r}^*, t^*, \psi^*$ & $\times$ & $\times$ & $\times$\\
$\mathbf{Em}_{\SI{2}{\meter}\times\SI{1}{\meter}}$ & $\times$ & {} & $\times$\\
$\mathbf{Em}_{\SI{6}{\meter}\times\SI{3}{\meter}}$ & {} & {} & $\times$\\
$\mathbf{Ls}_{1 ray/\SI{30}{\degree}}$ & {} & {} & $\times$\\
$4\times\mathbf{I}$ & {} & $\times$ & {}\\
\hline
\end{tabular}
\end{table}

% \begin{figure*}[ht]
% \includegraphics[width=\linewidth]{figures/terrains.png}
% \label{fig:terrains}
% \caption{Terrains. }
% \end{figure*}
% Should this be moved somewhere else or removed?
In this work, we design 9 basic terrains requiring different locomotion skills and train expert policies for each one. We then distill those skills into a single policy. During distillation, we modify the perceptive input. While the expert skills receive information about the terrain height around the robot's base in the form of an elevation map, the distilled policy must infer that information from the images of four onboard depth cameras. The policy must use its memory to infer parts of the obstacles that are not in the field of view of the cameras. Finally, we fine-tune the distilled policy using RL. During the fine-tuning stage, we extend the set of terrains by adding 3D scans of real-world search and rescue facilities representing piles of rubble, rocks, and collapsed buildings.

The remainder of this section describes each part of our proposed approach.

\subsection{Expert skill training}
\label{subsec:method/expert-skill-training}
% Basically, identical to ANYmal parkour + extra skills (thin wall, stepping-stones, rock pile, narrow beams), use symmetry (cite the latest paper)
The first step of our pipeline is to train the expert skills on individual terrains. We closely follow the training procedure used for the locomotion module of \cite{hoeller2023anymal}. We use the position-based task description proposed in \cite{rudin2022advanced} with symmetry data-augmentation of \cite{mittal2024symmetryconsiderationslearningtask}. On top of the five skills used by \cite{hoeller2023anymal} (walk, climb, climb down, jump, and crouch), we add four additional skills. These include jumping over low walls, walking on stepping stones, crossing narrow beams, and climbing piles of boulders. Each of the expert skills requires a specialized curriculum, reward-tuning, and training procedure. For example, the low-wall policy was initialized with the weights of the climbing policy. Finding the right procedure and tuning all parameters can be lengthy and cumbersome, but it only needs to be done once for each skill. Once sufficient performance is achieved, the trained policy can be stored and reused without further modifications or fine-tuning. 
% In this project, the five skills of \cite{hoeller2023anymal} were used without any modifications.
\subsection{Distillation}
\label{subsec:method/distillation}
\begin{figure*}[ht]
\centering
\includegraphics[width=0.9\linewidth]{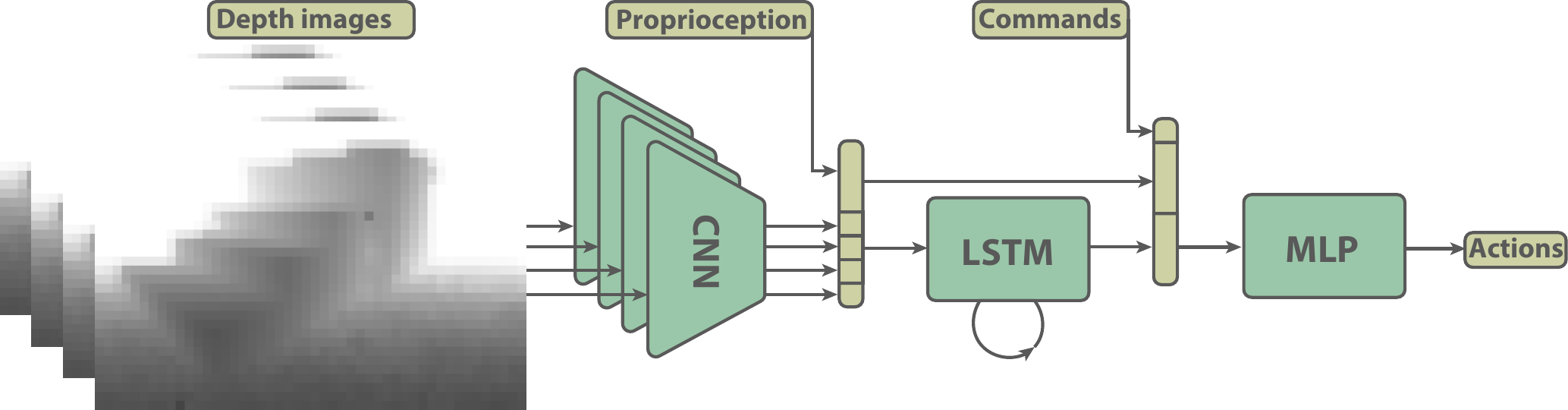}
\caption{Architecture of the policy used during distillation and fine-tuning. Three types of inputs are given to the policy: 4 depth images, proprioceptive information, and task commands. The depth images are processed individually by a CNN with 3 convolutional and max-pool layers, followed by two fully connected layers with an output dimension of 64. The features of the last fully connected layer of all images are then concatenated with the vector of proprioceptive information and fed through 2 LSTM layers. Finally, the output of the LSTM is concatenated with the vectors of proprioception and task commands and fed through an MLP composed of 3 fully connected layers with ELU activation.}
\label{fig:nn-architecture}
\end{figure*}

% terrains (additional randomization clutter), dagger, noisy-actions, observations, nn-architecture, loss
Once the different expert policies $\pi_{\text{expert}, i}$ are trained, they are distilled into a single foundation policy $\pi_{\text{student}}$. Formally this means that on every terrain $i$ given observations $o_{\text{student}}, o_\text{expert}$ the distilled policy should achieve $\pi_{\text{student}}(o_\text{student}) \approx \pi_{\text{expert}, i}(o_\text{expert})$. This highlights two abilities $\pi_\text{student}$ has to learn. First, the policy has to extract the type of terrain it currently encounters, i.e. it has to learn an internal mapping $o_\text{student} \mapsto i$. Otherwise, it would not be possible for $\pi_\text{student}$ to choose the right course of action. Then, it must predict the action chosen by that expert.

% If one of these two parts fails, then $\pi_\text{student}$ will not be able to overcome the current obstacle.  This challenging setting raises three questions to be addressed in this section:
% \begin{enumerate}
%     \item What observations $o_{\text{student}}$ should be chosen for $\pi_\text{student}$?
%     \item Which kind of network architecture should $\pi_\text{student}$ have?
%     \item How should $\pi_\text{student}$ be trained?
% \end{enumerate}

The distilled policy is trained online using supervised learning similar to the DAgger algorithm proposed in \cite{dagger}. We create a simulated environment that combines all terrains used for expert training. We then use the massively parallel simulation setting of \cite{rudin2022learning} and assign a subset of robots to each terrain. Each robot is then assigned an expert based on its terrain type. We collect trajectories by deploying $\pi_\text{student}$ in that environment. At each timestep, we query both the student and the assigned expert for each robot. The student's actions are sent to the simulation while the student's observations and the expert actions are collected in a dataset $\mathcal{D} = \{o_{\text{student}, t}, a_{\text{expert}, t}\}_{t=1}^T$. After collecting $\mathcal{D}$, the student policy is trained in a supervised fashion by minimizing the loss $\sum_{t=1}^N(\pi_\text{student}(o_{\text{student}, t}) - a_{\text{expert}, t})^2$. 

During the trajectory collection, zero-mean Gaussian noise is added to the actions. This leads to more robust distilled policies as it prevents overfitting to a small subset of trajectories. Additionally, having a student policy that is robust to action noise is beneficial for the RL fine-tuning stage, where such noise is added for exploration. The complete algorithm is outlined in Algorithm (\ref{alg:training}). 

% The environments used during distillation are similar to those used for expert training with further modifications to facilitate sim-to-real transfer (see Figure insert...).

\begin{algorithm}
\caption{Training Scheme for Policy Distillation}\label{alg:training}
\begin{algorithmic}
% \text{// assign each robot to an expert}
\State $\mathcal{E}_i \gets \textproc{getExpertIds()}$
\For{$k$ in $\leq N_{\text{Epochs}}$} 
    \State $\mathcal{D} \gets \emptyset$
    \For{$t \gets 0$ to  $T$} \Comment{Data Collection}
        \State $a_{\text{student}} \gets \pi_\text{student}(o_{\text{student}}) + \mathcal{N}(0, \sigma^2)$
        \For{$e \gets 0$ to $N_{\text{experts}}$}
            \State $a_{\text{expert}, \mathcal{I}[\mathcal{E}=e]} \gets \pi_\text{expert, e}(o_{\text{expert}})$
        \EndFor
        \State $\mathcal{D} \gets \mathcal{D} \cup \{(o_{\text{student}}, a_{\text{expert}})\}$
        \State $o_{\text{student}, t+1}, o_{\text{expert}, t+1} \gets$ \textproc{env.step}($a_{\text{student}}$)
    %     \State $t \gets t+1$
    \EndFor
    \State $\pi_\text{student} \gets \textproc{train}(\pi_\text{student}, \mathcal{D})$ \Comment{Supervised Training}
\EndFor
\end{algorithmic}
\end{algorithm}

The action space of the student policy is identical to the experts. 
All policies use proprioceptive information consisting of base linear and angular velocity, joint positions and velocities, the previous action, and the gravity vector projected into the base frame. Commands are given as target positions and orientations and a time to reach the target as proposed in \cite{rudin2022advanced}.

Expert policies use exteroceptive information in the form of an elevation map around the base. Due to the robustness and generalization issues described above, we instead use depth images for the student policy. We use 4 depth maps in total, two at the front of the robot and two at the back. 
% While the robot has depth cameras on the left and right sides, we find that adding them is not necessary to cross all obstacles successfully. On the other hand, not using side cameras reduces generalization problems when the robot walks next to walls, ledges, or humans.

We use a relatively simple network architecture for $\pi_\text{student}$ composed of 3 smaller building blocks. A convolutional network (CNN) processes each depth image individually to extract relevant exteroceptive information. Proprioceptive information is then combined with the exteroceptive features and used as input to a recurrent network (LSTM) (\cite{LSTM}). Finally, the LSTM's output is concatenated with the proprioceptive information and commands and fed through a fully connected network (MLP), which predicts the next action. The network architecture is visualized in Figure \ref{fig:nn-architecture}.

\subsection{RL-finetuning}
\label{subsec:method/rl-finetuning}
% terrains, rewards, observations, terminations, curriculum, tricks for stability (critic-pretraining, low learning rate, cite papers claiming that RL-finetuning is hard)
After distillation, the student policy is capable of imitating the actions of all experts. However, as can be seen in Table \ref{table:success-rate}, the performance on individual skills decreases, and the generalization to unseen terrains is low. The problems lie in the fact that during distillation, the policy is not directly trained to solve the task, but rather, guess which of the experts it should imitate and what that expert would do. We propose to improve the performance of the distilled policy using RL. On its own, RL is not suitable for training a policy for all terrains from scratch. Due to inefficient exploration, the policy collapses to a single mode and only learns to solve a subset of the different tasks or terrains. However, if a policy reaches sufficient performance through distillation, it can then be fine-tuned with RL leading to an increased performance across all tasks. 
Unfortunately, training a distilled foundation policy with supervised learning and then fine-tuning it with RL tends to be unstable. Done naively, the performance of the policy steadily degrades once the RL loss is applied. This problem can be circumvented using three key components: maximum performance and robustness of the foundation policy under the action noise added during fine-tuning, conservative tuning of the RL hyper-parameters, and efficient pre-training of the critic network.

We achieve robustness by adding action noise during the distillation process and reducing the initial standard deviation of the RL policy distribution. In terms of the precision of a critic network, we find that we need to pre-train the critic network to a sufficient level before starting to update the policy. We achieve this by having an initial phase of the RL training where the policy weights are frozen, and hyper-parameters are tuned to maximize the training efficiency of the critic. 
% Note that we could not achieve the same result by pre-training the critic network during distillation. This can be caused by the different time horizons used during distillation and/or the environmental modifications added for fine-tuning. After this initial phase, the standard RL training resumes with hyper-parameters restored to the ones used in our previous work, except the learning rate, which needs to remain an order of magnitude lower for stable training.

% rewards/terminations/terrains

% Additionally, for efficient fine-tuning, the policy must choose sensible motions on unseen terrains to reduce the exploration problem. 

\subsection{Depth image noise-model}
\label{subsec:method/depth-noise-model}
\begin{figure*}[ht]
\includegraphics[width=\linewidth]{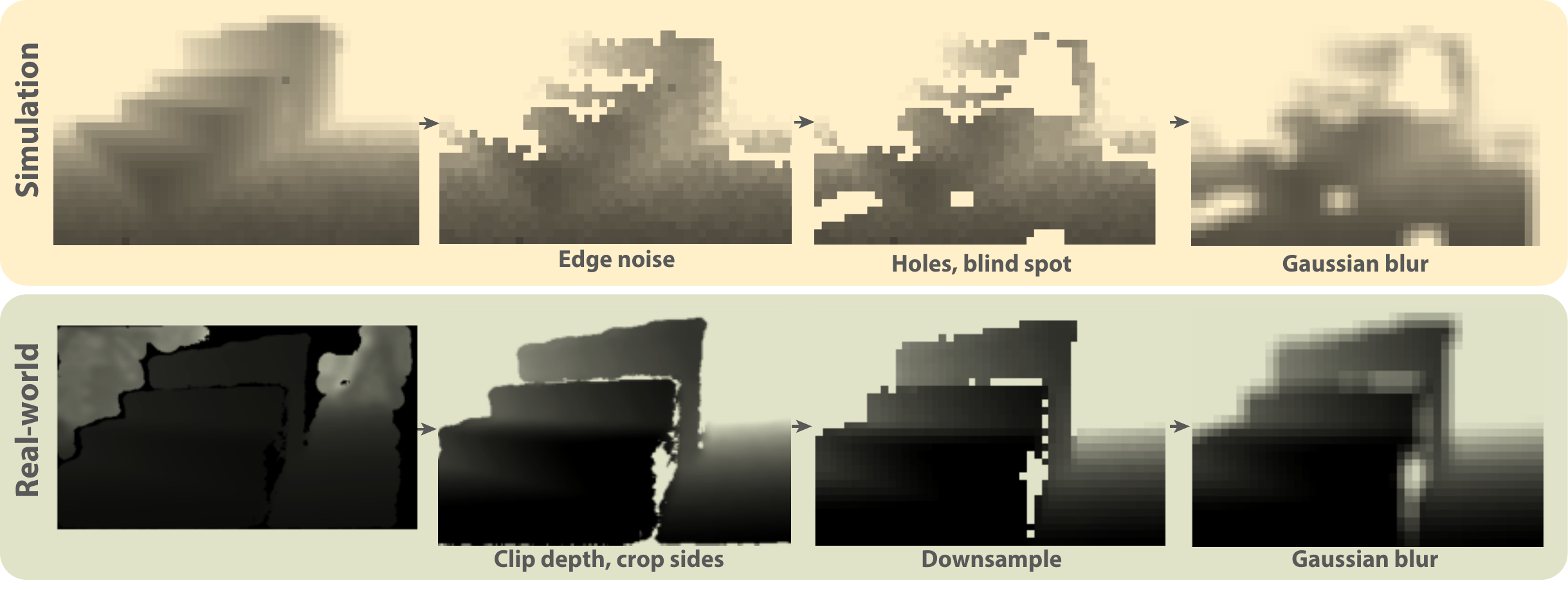}
\caption{Processing applied to simulated and real depth images. Simulated images are degraded using the following steps: 1) pixels surrounding edges are shuffled and/or removed, 2) random holes are added using slowly evolving Perlin noise, and 3) the image is blurred using a Gaussian filter. Real images are 4) clipped, downsampled, and cropped, and 5) blurred using the same Gaussian filter as in simulation.}
\label{fig:depth-noise-model}
\end{figure*}
% sim : edge removal, noise, pixel shuffling, holes based on perlin noise, removal of left side of image (stereo matching blind spot), blurring. Real: downscaling, cut sides, blur.
Training policies directly from depth images create the challenge of sim-to-real transfer for those images. Specifically, the characteristics of the ANYmal's Realsense D435i cameras need to be replicated in simulation. These cameras rely on stereo-matching as well as infrared projection to compute depth information. These processes have limitations that lead to imperfections in the images. These limitations include missing data due to a lack of stereo-matching around the edges of objects, on some surfaces, for objects too close to the camera, and depending on the distance on the left or right side of the image. Additionally, the depth precision is degraded at larger distances.

While getting ground truth depth information from the simulation is relatively straightforward, properly simulating the stereo-matching process is not practical. To bridge the gap, we process both real and simulated images as shown in Fig. \ref{fig:depth-noise-model}. In simulation, we begin by rendering a low-resolution image of 48x32 pixels. We then degrade the image using the following steps:
\begin{enumerate}
\item \textit{Clip}: The depth is clipped at \SI{2}{m}. Additionally, pixels with a value below \SI{0.15}{m} are considered empty (set to \SI{2}{m} as well).
\item \textit{Edge noise}: we identify edges by thresholding the depth gradient across the image. Pixels around the edges are then randomly set to empty or shuffled with neighboring pixels.
\item \textit{Holes}: Patches of the image are set to the maximum depth. The patches are identified by thresholding slowly evolving Perlin noise \cite{PerlinNoise}, which ensures temporal consistency of the missing data.
\item \textit{Blind spot}: we remove data in the leftmost 1 to 5 columns of the image. This mimics the blind spot of stereo-matching at small distances.
\item \textit{Gaussian Blur}: we blur the whole image using a Gaussian kernel, which removes minor details and further helps bridge the sim-to-real gap.
\end{enumerate}

We apply simple processing steps to real-world images to match the characteristics of the simulated ones. We clip the depth the same way as for simulated images, down-sample to the correct size, and finally apply the same Gaussian blur as in the simulation.
\section{Experiments}
\label{sec:experiments}
\subsection{Success rate on different terrains}
\label{subsec:experiments/success-rate}

\begin{table*}[t]
\small\sf\centering
\caption{Success rate of policies on different terrains. $\pi_{w}$ to $\pi_{ss}$ are the experts trained on \textit{Walk}, \textit{Climb}, \textit{Climb down}, \textit{Jump}, \textit{Tables}, \textit{Rock pile}, \textit{Low wall}, \textit{Beams}, and \textit{Stepping stones}, respectively. $\pi_{D}$ and $\pi_{RL}$ are the distilled and fine-tuned policies. Terrains from \textit{Walk} to \textit{Stepping stones} are used during distillation. \textit{Parkour line} and \textit{Scanned meshes (train)} terrains are added during fine-tuning. The other terrains are never seen during training and are used only to evaluate the generalization capabilities of policies. $\pi_{RL^*}$ is fine-tuned again on the \textit{Climb down on stones}. The highest success rate across policies and all rates within \SI{0.5}{\%} are in bold.}
\label{table:success-rate}
\begin{tabular}{lccccccccc|ccc}
\toprule
Terrain & $\pi_{w}$ & $\pi_{c}$ & $\pi_{cd}$ & $\pi_{j}$ & $\pi_{t}$ & $\pi_{rp}$ & $\pi_{lw}$ & $\pi_{b}$ & $\pi_{ss}$ & $\pi_{D}$ & $\pi_{RL}$ & $\pi_{RL^*}$\\
\midrule
\texttt{Walk} & 94.6 & 54.4 & 36.7 & 43.8 & 23.4 & 96.4 & 30.4 & 25.4 & 33.2 & 99.3 & \textbf{100.0} & \textbf{99.8} \\
\texttt{Climb} & 0.0 & 98.8 & 0.0 & 0.0 & 0.0 & 0.0 & 0.0 & 0.0 & 0.0 & 98.1 & \textbf{99.5} & \textbf{99.4}\\
\texttt{Climb Down} & 2.6 & 13.4 & \textbf{99.9} & 47.6 & 3.1 & 43.0 & 2.1 & 8.8 & 6.0 & 84.3 & \textbf{99.7} & \textbf{99.6}\\
\texttt{Jump} & 0.7 & 2.9 & 0.0 & \textbf{98.4} & 0.0 & 9.7 & 0.0 & 0.0 & 0.0 & 93.5 & \textbf{98.0} & 97.7\\
\texttt{Tables} & 0.2 & 24.4 & 0.0 & 0.0 & 99.4 & 37.2 & 0.0 & 0.0 & 0.0 & 78.9 & \textbf{100.0} & \textbf{100.0}\\
\texttt{Rock pile} & 34.6 & 14.6 & 31.5 & 25.9 & 2.6 & 92.2 & 24.9 & 4.0 & 12.4 & 82.6 & 96.5 & \textbf{97.1}\\
\texttt{Low wall} & 0.0 & 14.8 & 0.0 & 0.0 & 0.0 & 21.8 & 84.8 & 0.0 & 0.0 & 77.0 & \textbf{99.9} & \textbf{100.0}\\
\texttt{Beams} & 0.3 & 3.2 & 0.0 & 8.7 & 0.3 & 0.4 & 0.1 & 97.3 & 0.3 & 85.2 & \textbf{99.5} & \textbf{99.5}\\
\texttt{Stepping stones} & 1.1 & 1.3 & 0.0 & 23.4 & 0.0 & 0.0 & 0.0 & 1.0 & \textbf{98.8} & 73.0 & \textbf{98.8} & \textbf{98.9}\\
\midrule
\texttt{Parkour line} & 0.2 & 0.0 & 0.1 & 0.2 & 18.6 & 17.2 & 0.0 & 0.0 & 0.0 & 5.8 & \textbf{98.5} & \textbf{98.7}\\
\texttt{Scanned meshes (train)} & 0.2 & 1.6 & 0.0 & 0.0 & 0.0 & 44.8 & 0.0 & 0.0 & 0.0 & 11.9 & \textbf{99.1} & \textbf{98.8}\\
\midrule
\texttt{Scanned meshes (test)} & 0.2 & 1.6 & 0.0 & 0.0 & 0.0 & 44.8 & 0.0 & 0.0 & 0.0 & 14.9 & \textbf{94.9} & 93.9\\
\texttt{Arranged rocks} & 31.1 & 55.6 & 17.2 & 17.0 & 6.8 & 67.7 & 13.7 & 13.0 & 9.7 & 62.9 & \textbf{93.2} & \textbf{92.8}\\
\texttt{Gap - climb} & 0.0 & 0.0 & 0.0 & 0.0 & 0.0 & 0.0 & 0.0 & 0.0 & 0.0 & 10.2 & 82.0 & \textbf{86.6}\\
\texttt{Down - stones} & 0.0 & 0.0 & 0.0 & 17.8 & 0.0 & 0.0 & 0.0 & 0.0 & 1.3 & 11.3 & 54.4 & \textbf{92.4}\\

\bottomrule
\end{tabular}\\[10pt]
\end{table*}

First, we examine the performance of different policies across various terrains, both seen and unseen during training. In particular, we compare the performance of individual skills, the distilled policy, and finally, the fine-tuned policy. The experiment is done on the following terrains:
\begin{itemize}
    \item Terrains for expert training and distillation
    \begin{itemize}
        \item The nine terrains with individual experts.
    \end{itemize}
    \item Terrains added during RL Fine-tuning
    \begin{itemize}
        \item \textit{Parkour line} - a mix of boxes, gaps, tables, stairs, and slopes, identical to one of the terrains used in \cite{hoeller2023anymal}.
        \item \textit{Scanned meshes (train)} - real-world scans of search and rescue training facilities used during fine-tuning.
    \end{itemize}
    \item Unseen terrains
    \begin{itemize}
        \item \textit{Scanned meshes (test)} - real-world scans that were not used during fine-tuning.
        \item \textit{Arranged rocks} - manual arrangement of rocks requiring motions similar to expert skills.
        \item \textit{Gap - climb} - elevated platform separated by a gap, requiring a mix of the \textit{Climb up/down} and \textit{Jump} skills.
        \item \textit{Climb down on stones} - elevated platform separated surrounded by stepping stones, requiring a mix of the \textit{Climb down} and \textit{Stepping stones} skills.
    \end{itemize} 
\end{itemize} 
Table \ref{table:success-rate} shows the average success rate computed by collecting 1000 roll-outs in simulation on randomized terrains at 90\% of the maximum training difficulty. The results of the experiments provide insights into both the distillation and fine-tuning processes. 

After distillation, the resulting policy is capable of imitating all expert skills, but its performance is uneven. On average, there is \SI{10.4}{\%} drop in success rate, but we see a larger decrease on terrains with ambiguity between skills, such as \textit{Tables} versus \textit{Climb} and \textit{Climb Down} versus \textit{Jump}. In those cases, the distilled policy does not always succeed at identifying the terrain and tends to learn a sub-optimal mix of experts' behaviors. Additionally, we see a significant decrease in terrains requiring precision, such as \textit{Stepping stones} and \textit{Beams}, where the policy may not be able to imitate the expert due to the change of the perceptive modality. Interestingly, the distilled policy performs better than the expert skill on the standard \textit{Walk} terrains. This shows that there is some level of knowledge reuse between skills that naturally emerge from the distillation. By learning complex motions on other terrains, the policy naturally starts using them and manages to cross some of the obstacles that the expert skill could not.
Looking at terrains not present during distillation, we see a low success rate overall, which shows that despite being able to imitate nine different experts, the distilled policy does not generalize to unseen terrains.

After fine-tuning, the average performance increases significantly and matches or even surpasses the success rate of all nine experts. On average, it has a success rate 3.1\% higher than the corresponding expert with a significant increase on the \textit{Low wall} terrain.
As expected, the success rate is also significantly higher for the \textit{Parkour line} and \textit{Scanned meshes} added during the fine-tuning stage. Finally, we get mixed results on terrains not seen during any of the training stages. While we see good generalization to unseen meshes, arranged rocks, and the \textit{Gap - climb} terrain, the combination of motion in \textit{Climb down on stones} proves to be a significant challenge for policies even with a lower platform and larger stones compared to the \textit{Climb down} and \textit{Stepping stones} terrains. This forces us to conclude that while we see signs of generalization emerging from our method, the policy can still fail on out-of-distribution terrains, even in relatively simpler scenarios.

\subsection{Repeated Fine-tuning on new terrains}
\label{subsec:experiments/skill-combination-methods}
\begin{figure}[b]
\centering
\includegraphics[width=0.8\linewidth]{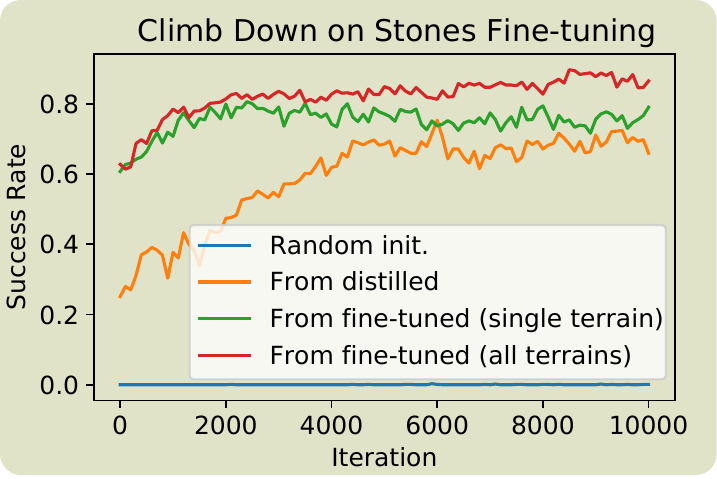}
\caption{Reconstruction loss during distillation for two observation modalities (elevation map and depth images) and two network architectures with and without memory (LSTM and MLP, respectively). Average and standard deviation over 8 runs.}
\label{fig:fine-tuning-plot}
\end{figure}
Our method allows us to keep fine-tuning the policy on new terrains. This allows us to continually extend the capabilities of the policy when, during testing, we encounter a new type of obstacle on which the performance of the policy is unsatisfactory. To test the viability of the repeated fine-tuning approach, we select a terrain that none of the policies were capable of solving successfully and analyze the capacity of different policies to adapt to that new obstacle type when fine-tuned for it. Based on the results of Table \ref{table:success-rate}, we select the \textit{Climb down on stones} terrain for this experiment. We compare training a policy from scratch, fine-tuning from the distilled policy, and repeated fine-tuning from the policy that was already fine-tuned with RL on other terrains. For the repeated fine-tuning approach, we additionally compare training on only the new terrain with training on a combination of all of the previously used terrains extended with the new one. In that case, the new terrain represents only \SI{3}{\%} of the collected samples.

Figure \ref{fig:fine-tuning-plot} shows the success rate of the different policies during training. We see that a policy trained from scratch on this terrain does not find a suitable solution. The prior contained in the distilled policy is sufficient to start training and leads to non-trivial performance. However, starting the training from the already fine-tuned policy leads to both faster training and higher final performance. Interestingly, training on all terrains leads to better final performance than using only the new terrain, even though, in that case, only \SI{3}{\%} of samples are collected on the new terrain. Keeping a diversity of training terrains seems important to achieve the highest performance on new and complex obstacles. Additionally, we show the success rate of the resulting policy on other terrains in the last column of Table \ref{table:success-rate}. We see that the performance does not change significantly on previously seen terrains while increasing on the newly added one, which shows that continual fine-tuning is a viable approach to adding new capabilities to the policy. We note that more analysis is needed to understand the limits of this method. While it allows us to train a policy on tens of different terrains, we expect to see decreasing performance when more terrains requiring specialized motions are added to the training.

\subsection{Skill combination methods}
\label{subsec:experiments/skill-combination-methods}
\begin{figure*}[t]
\includegraphics[width=\linewidth]{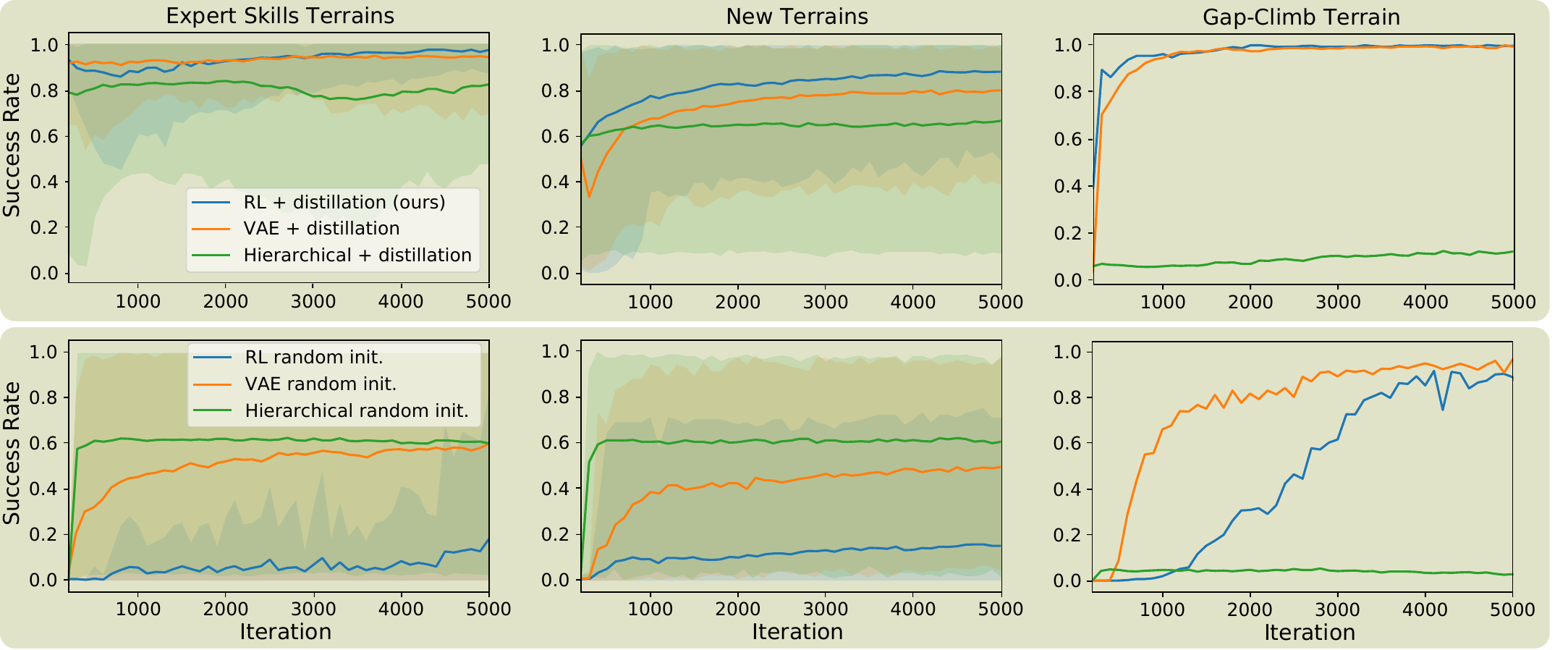}
\caption{Comparison of three skill combination methods. We compare our proposed distillation + fine-tuning approach to a hierarchical approach where a policy is trained to switch between experts and a method in which we use a VAE to encode motions of skills into a latent space, freeze the decoder and then train a new policy controlling the robot through the resulting latent space. The top row shows the results for pre-trained policies. Both the hierarchical and VAE policies can be trained from scratch or pre-trained in the setup used for distillation by adapting the desired output (one hot encoding and latent vector, respectively). The bottom row shows the results without pre-training (Note that our approach without pre-training is standard RL training from scratch). We evaluate the performance using 1000 robots on 100 randomized terrains with difficulties from \SI{50}{\%} to \SI{100}{\%} of the maximum training difficulty. We use the mean and mode of the policies. The solid curve shows the mean success rate across all 100 terrains. The shaded areas show the minimum and maximum success rates across terrain types.}
\label{fig:training-plots}
\end{figure*}
We compare our proposed approach with other methods used in previous work to combine separate motor skills. The first baseline is the hierarchical approach of \cite{hoeller2023anymal}, where a policy is trained to select and command one of the expert skills. The high-level policy can choose a different skill and change its command at every timestep. Second, we develop an approach similar to \cite{bohez2022imitate, luo2024universal}. We use a variational autoencoder (VAE) to embed motions of the different skills into a common latent space. Then, the decoder is frozen, and a new policy is trained to control the robot through that latent space. The purpose of this approach is to create a new action space that simplifies the exploration problem and is potentially better suited for interpolation between skills.
For all approaches, we compare training from scratch as well as starting from a network pre-trained in the distillation environment. Note that our approach without pre-training is standard end-to-end RL. 
Figure \ref{fig:training-plots} shows the results of the comparison on three different sets of terrains, including the terrains of expert skills used during distillation, a combination of new terrains added during fine-tuning, and the \textit{Gap-Climb} terrain mentioned above which is designed to test interpolation between the climbing and jumping skills. 

The results show that our proposed distillation and fine-tuning approach outperforms all others. Focusing first on pre-trained networks (second row of the figure), we can see that both our and VAE approaches perform well across all sets of terrains. On new terrains, the distillation and fine-tuning approach performs better. This can be explained by the fact that new motions required by those terrains are harder to achieve for the VAE since the decoder was trained only on the motions of the skills.
The hierarchical method does not reach satisfactory performance. It struggles even on terrains with expert skills where it only needs to select the correct skills and forward the command. It tends to perform well on most of the skills but completely stops using some of them, leading to constant failure on corresponding terrains. Additionally, it completely fails on the \textit{Gap-Climb}. This is expected as this terrain requires motions combining two skills, which is not possible with the hard-switching of the hierarchical approach.

For networks trained from scratch (1st row of the figure), we can see a clear advantage for the VAE over both other methods. We see that exploration in the latent space indeed leads to more efficient training.
The hierarchical policy quickly reaches the same performance level as its pre-trained version but does not outperform it. Standard RL with random initialization takes a long time to learn anything and does not manage to find solutions for all terrains. Similar to the hierarchical approach it manages to solve a sub-set of the terrains while remaining completely stuck on others. Note that this comparison is not completely fair since the Standard RL method requires no prior training, while both the VAE and the hierarchical approach require training expert skills. The VAE additionally requires training the decoder before the RL stage begins.

From these results, we can conclude that the hierarchical approach and RL without distillation are not viable methods for solving a large variety of terrains or, more generally, tasks with a single controller. As shown by prior work, both methods can be used successfully for a reduced set of tasks, but our results indicate that they lack scalability when more skills or tasks are added.

Both distillation with fine-tuning or VAE latent space encoding with re-training seem to be possible approaches. However, given the slightly higher performance of the distillation approach on new terrains and the additional complexity of the VAE approach, we select distillation with RL fine-tuning as our proposed method.

\subsection{Effect of network architecture}
\label{subsec:experiments/effect-of-nn-architecture}
\begin{figure}[b]
\centering
\includegraphics[width=0.8\linewidth]{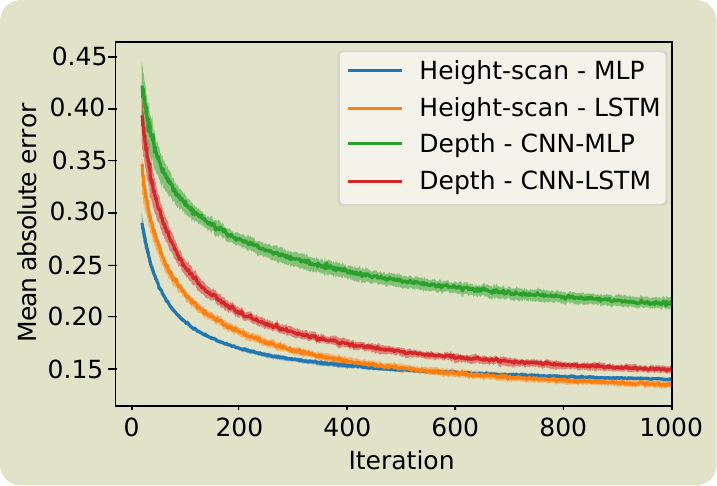}
\caption{Reconstruction loss during distillation for two observation modalities (elevation map and depth images) and two network architectures with and without memory (LSTM and MLP, respectively). Average and standard deviation over 8 runs.}
\label{fig:distillation-plot}
\end{figure}
% Success rate during training of from-scratch, from-distilled, vae-from-scratch, vae-from-distilled, hierarchical-from-scratch, hierarchical-from-distilled [3 plots: expert terrains, 1 new terrain (pallets), all other new terrains]
In the next experiment, we analyze the necessity of having a memory mechanism in the neural network architecture when using different observation modalities. In particular, we compare the distillation performance of a neural network with and without a recurrent LSTM component. The LSTM version is identical to the architecture described in Section \ref{subsec:method/distillation}. The MLP version simply removes the LSTM and directly feeds the CNN features to the MLP.

We use two types of perceptive observations: depth images and elevation maps. The elevation maps contain a combination of small fine-grained (1x2m at 0.1m resolution) and larger coarse (3x6m at 0.5m resolution) terrain data. We add a horizontal, Lidar-like scan, allowing the policy to distinguish overhanging obstacles. The depth images come from the two forward and two backward-facing onboard cameras of the robot. They are processed with the noise model described above.

Figure \ref{fig:distillation-plot} shows the mean absolute error for the two architectures with the two observation types. The results indicate that while both architectures perform well with the elevation map observations, a memory component is needed to extract useful information from the depth images. Additionally, we see that even with memory the error remains higher when the depth images are used. This can be explained by the fact that in some cases the distilled policy does not have enough information to distinguish the terrains, for example, thin walls and boxes are indistinguishable until the robot starts climbing.

\subsection{Active perception}
\begin{figure}[h]
\centering
\begin{subfigure}{.48\linewidth}
  \includegraphics[width=\linewidth, trim={10cm 3cm 15cm 5cm},clip]{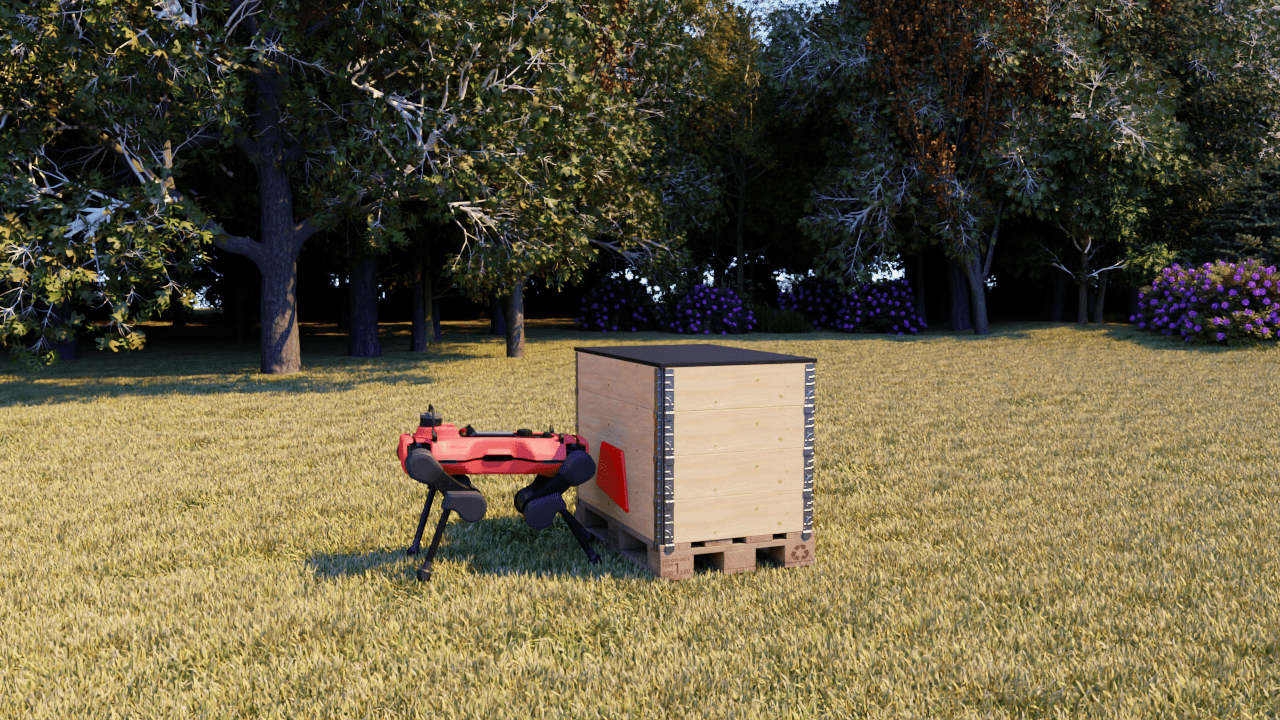}
  \includegraphics[width=\linewidth, trim={10cm 3cm 15cm 5cm},clip]{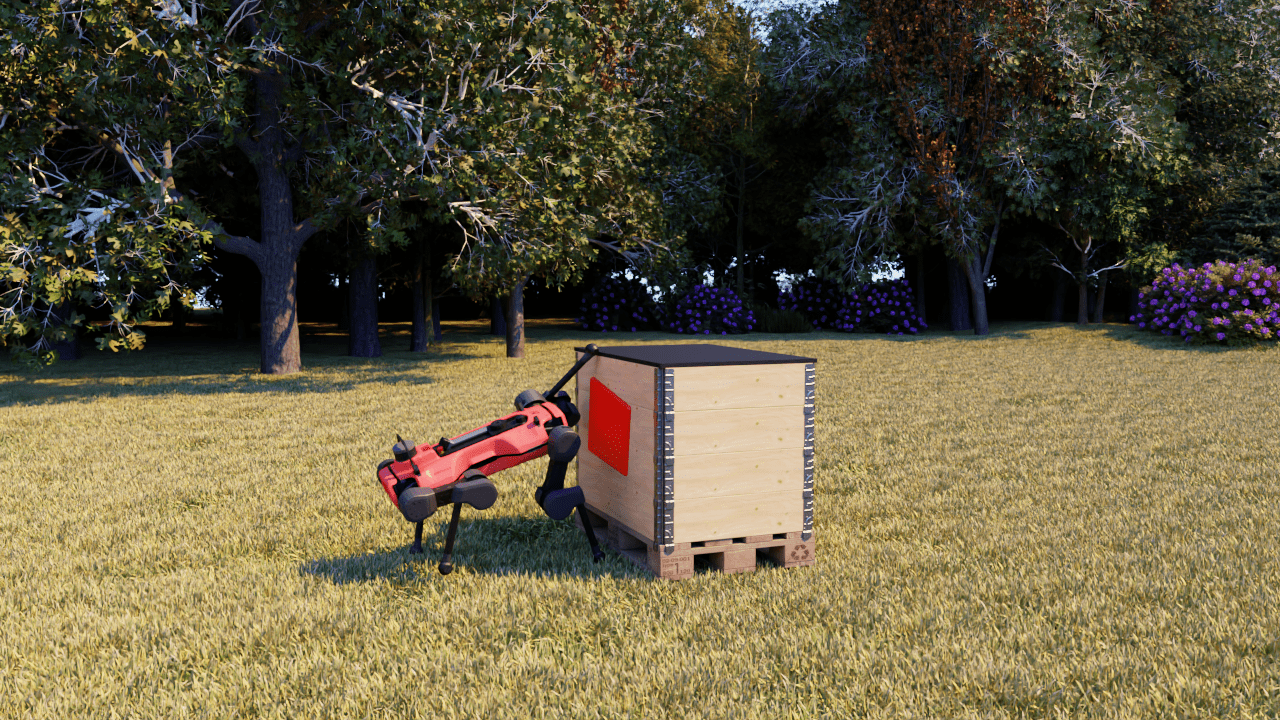}
  \caption{After distillation}
  % \label{fig:sfig1}
\end{subfigure}%
\hspace{0.01\linewidth}
\begin{subfigure}{.48\linewidth}
  \includegraphics[width=\linewidth, trim={10cm 3cm 15cm 5cm},clip]{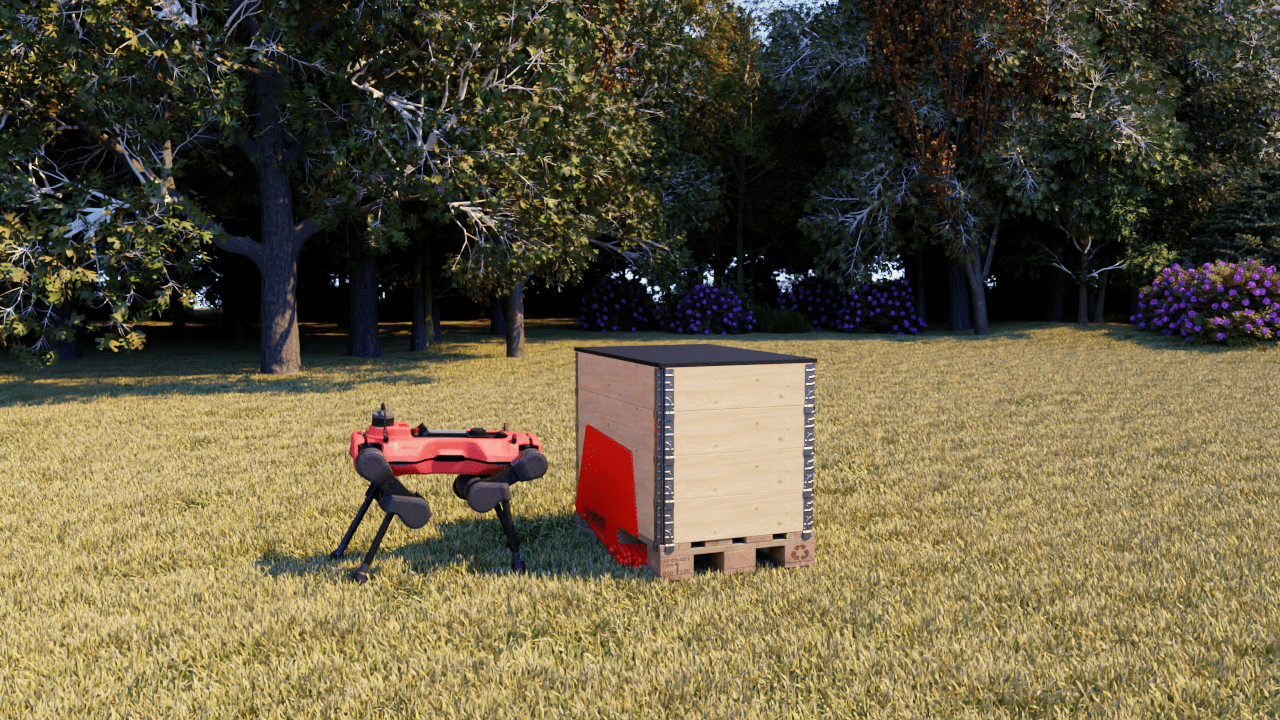}
  \includegraphics[width=\linewidth, trim={10cm 3cm 15cm 5cm},clip]{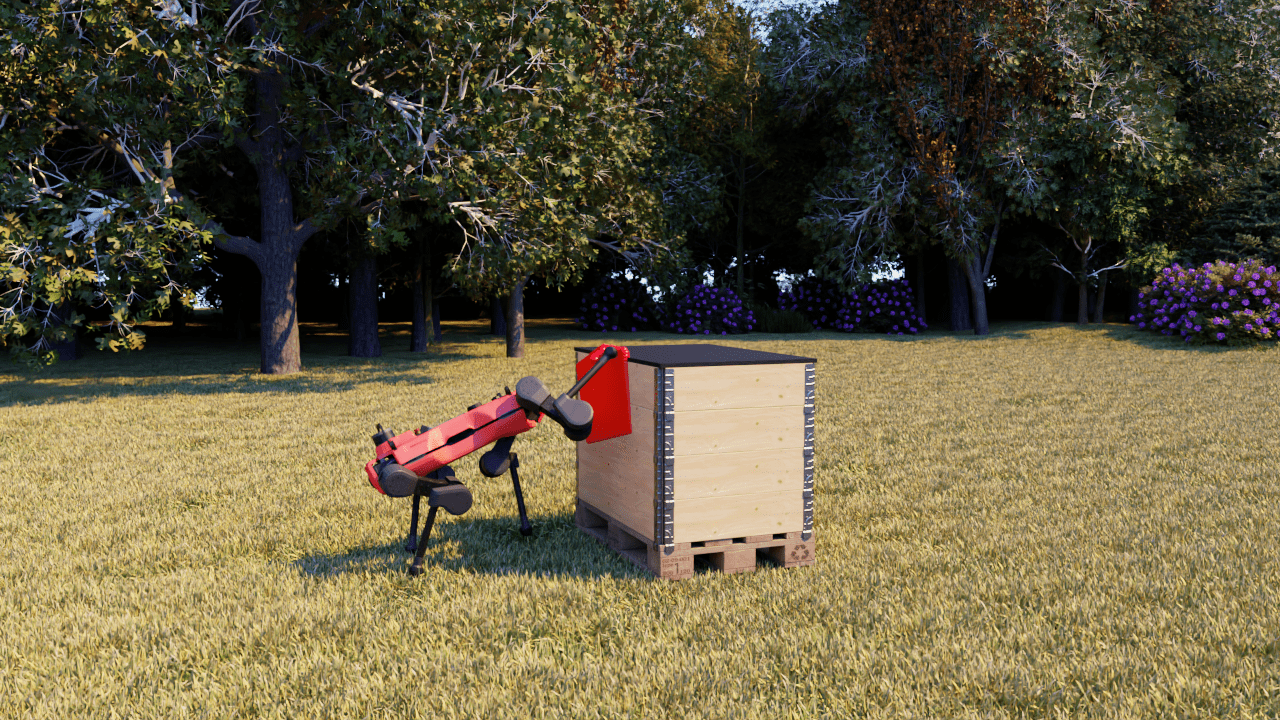}
  \caption{After fine-tuning}
  % \label{fig:sfig2}
\end{subfigure}
\caption{Change of behavior between the distilled and fine-tuned policy. After fine-tuning, the policy learns to stay further away from the obstacle to maximize the visibility of the obstacle in the depth camera's field of view.}
\label{fig:active-perception}
\end{figure}
Finally, we show that our approach allows the policy to learn new behaviors enhancing its perceptive capabilities. Our expert skills are trained with nearly perfect elevation maps. The information contained in these maps is independent of the pitch and roll of the robot and is not altered by occlusions. On the other hand, depth cameras are rigidly fixed to the robot, they follow the full pose, and due to their position have a reduced view of the obstacles. 

During distillation, the student policy is trained to imitate the motion of the expert without considering that this motion might be sub-optimal under its perception modality. However, once we fine-tune the policy with RL, it can learn to modify its behavior to increase the success rate of the motion.

Figure \ref{fig:active-perception} shows an example of such adaptation. The climbing expert skill learns to stop very close to the box it is about to climb. While this reduces torques during the stand-up phase, it significantly reduces the field of view of the depth cameras. We can see that the robot reaches for the top of the box without seeing it. Performing this motion is possible by memorizing the image of the obstacle seen during the approach, but it results in lower reliability and robustness to disturbances. After fine-tuning, the policy learns to stop further away from the obstacle and tilts the body of the robot in a way that brings the top of the box within the field of view before the leg touches it. This behavior is consistent across different training runs and obstacle shapes. Similarly, the policy changes its approach behavior for other obstacles, such as stepping stones or tables, but the benefits of those behaviors are less obvious to the human eye.

\subsection{Real-world deployment}
\label{subsec:experiments-real-world-deployment}
\begin{figure*}[t]
\centering
\begin{subfigure}{.49\linewidth}
  \includegraphics[width=\linewidth]{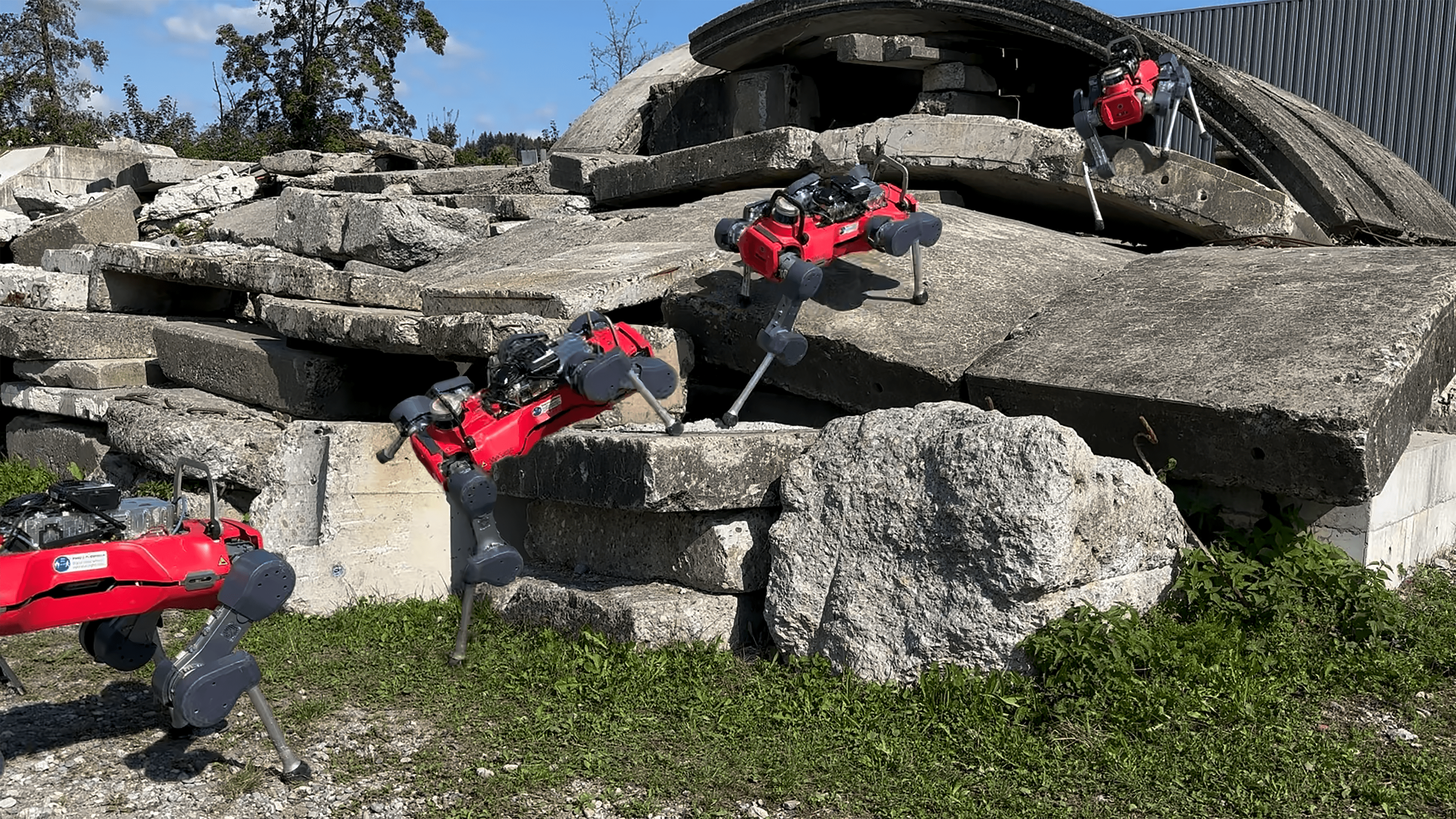}
  \includegraphics[width=\linewidth]{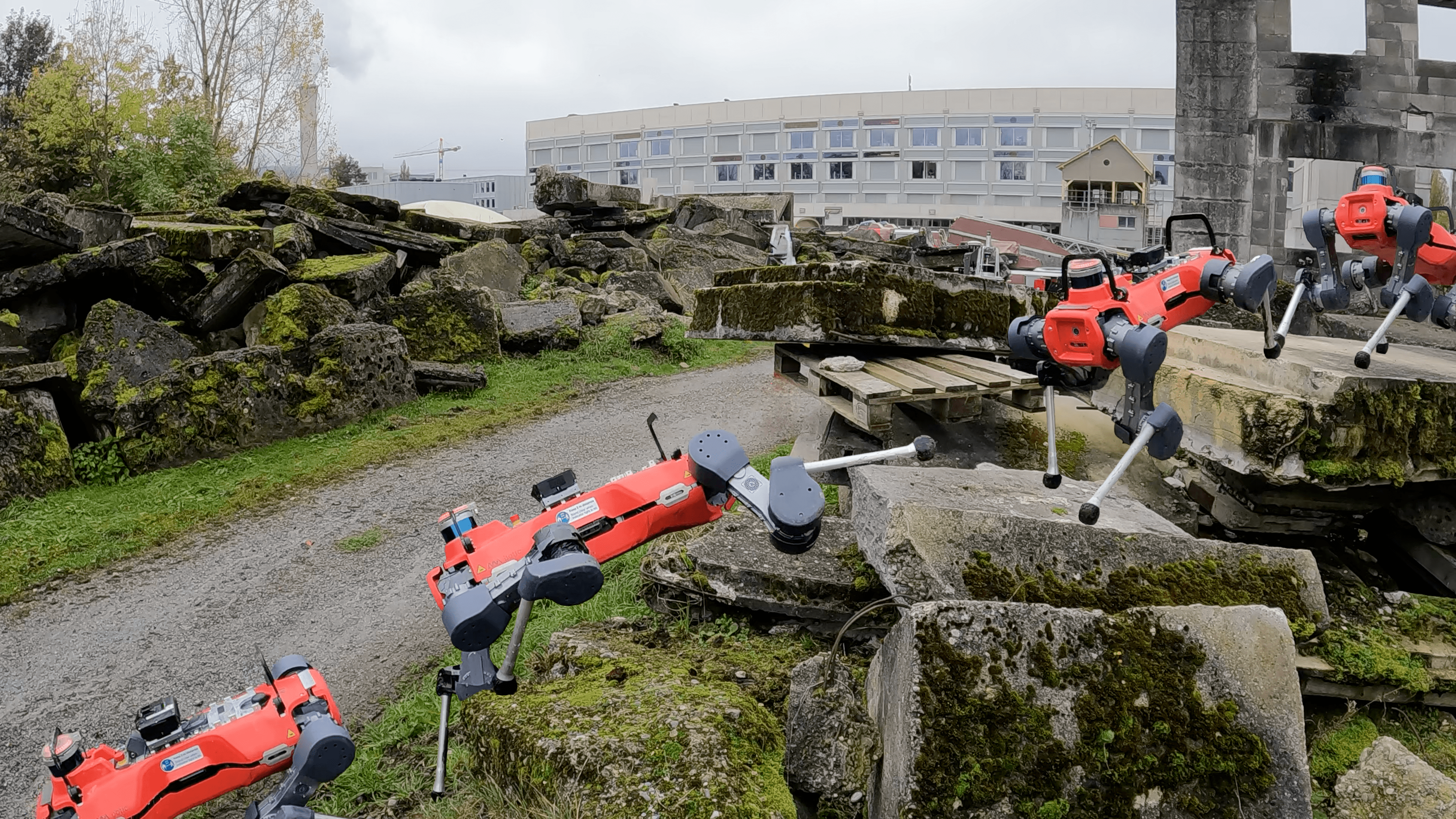}
  \caption{Search and rescue facility.}
  % \label{fig:sfig1}
\end{subfigure}%
\begin{subfigure}{.49\linewidth}
  \includegraphics[width=\linewidth]{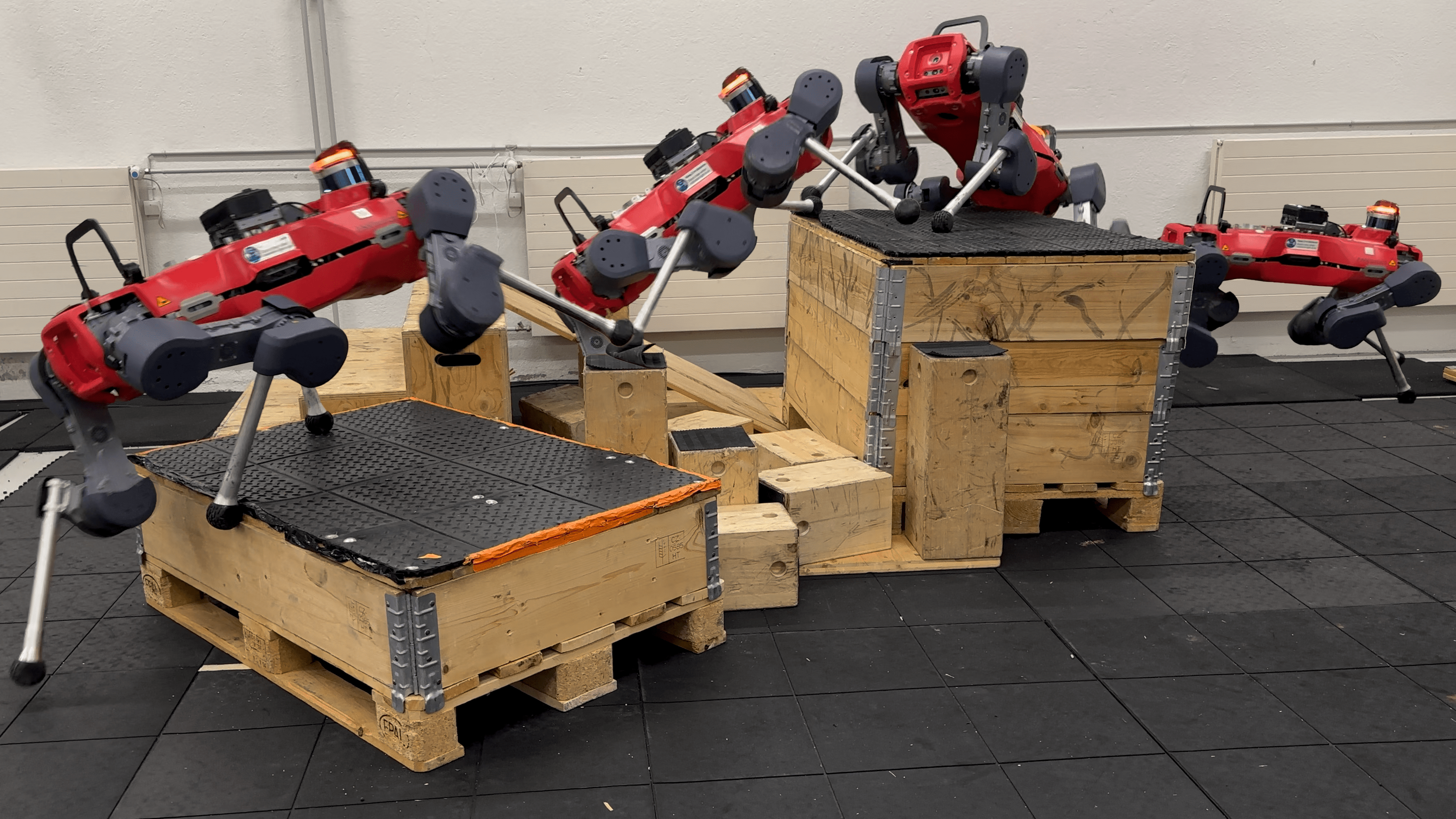}
  \includegraphics[width=\linewidth]{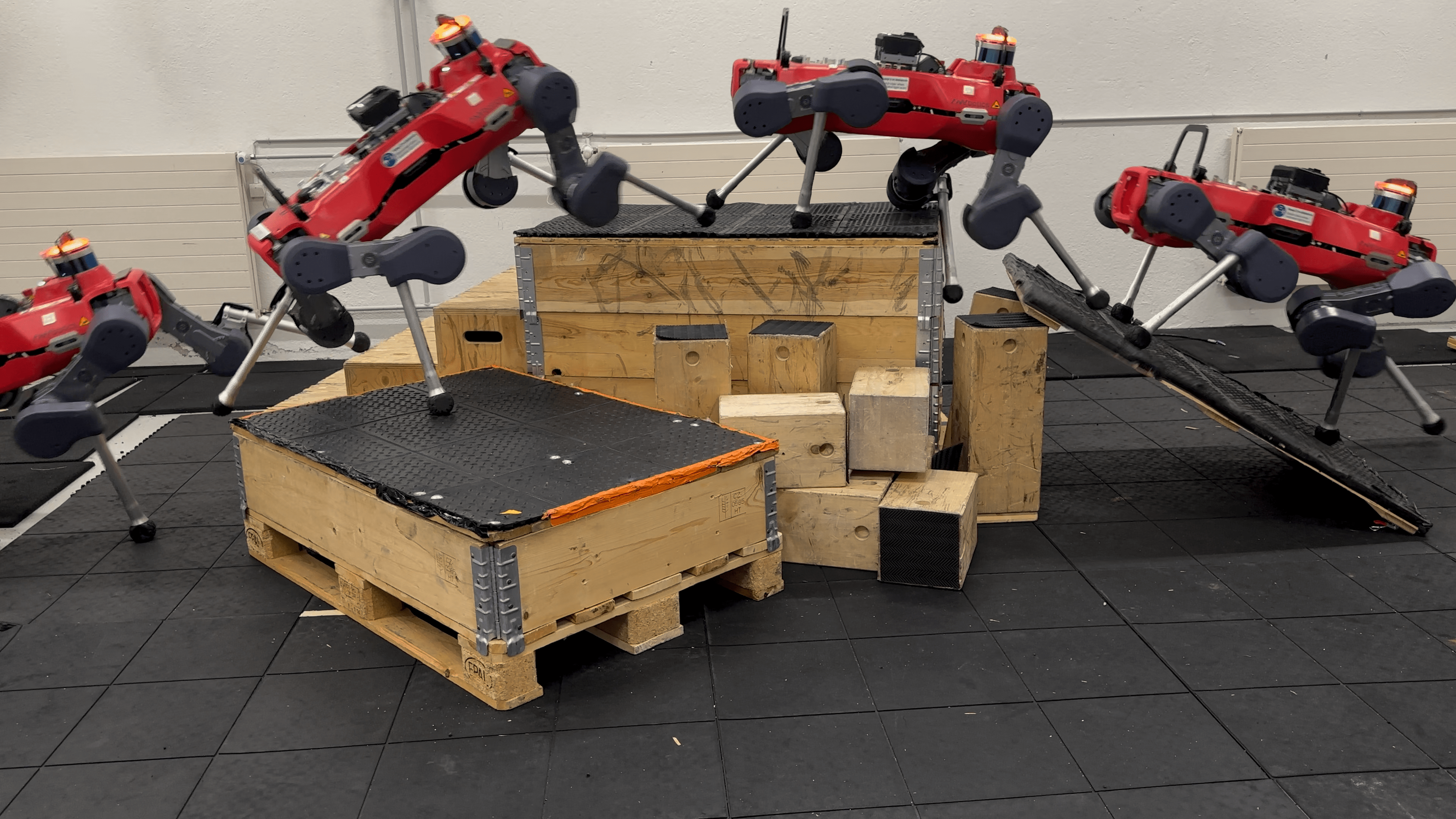}
  \caption{Indoor}
  % \label{fig:sfig2}
\end{subfigure}
\caption{Real-world deployment of the ANYmal D robot controlled by the fine-tuned policy. Both indoor and outdoor terrains are not seen during training.}
\label{fig:deployment}
\end{figure*}

After fine-tuning, the policy is ready to be deployed in the real world. We deploy the policy on a standard ANYmal D robot. Our policy uses four out of the six available depth cameras and does not use the onboard Lidar or any external sensors. All computations happen on the onboard CPU at 50Hz. The depth images are updated at 15Hz and processed as described in Section \ref{subsec:method/depth-noise-model}. Internally, the cameras update and send images as ROS messages at 60Hz. Since we randomize the delay of each camera during training, we do not need to use any synchronization mechanism during deployment. We simply provide the latest received image for each camera.

We deploy the robot in various indoor and outdoor environments. Indoor, we show that the policy can execute the correct skills across isolated obstacles as well as handle an unstructured collection of obstacles. Outdoors, we test our policy in search and rescue training facilities on piles of rubble imitating collapsed buildings. The terrains used during these tests were not seen during training.
The supplementary video shows these deployments. The robot can successfully navigate different environments.
 The robot executes dynamic motions such as climbing and jumping while adapting the motion to unstructured terrains. The policy shows exceptional robustness to various disturbances, including slippery or moving ground, feet being caught in cracks or steel wire, and visual distractors such as tall grass or degraded depth due to reflective surfaces and direct sunlight. 
 
 On the other hand, the behavior, and more specifically, the precision of the policy, can still be improved. The footstep selection is not perfect. Even though the robot recovers quickly, it could easily avoid some of the missteps. Additionally, during climbs and jumps, the policy learns to use the knees of the robot more than the expert skills. This is understandable since it is a safer approach compared to trying to land on the feet directly, but it also leads to increased impacts on the motors and, thus, faster wear of the hardware.

\section{Discussion}
\subsection{Skill combination methods}
% From our experiments, we can conclude that our proposed approach is suitable for combining separately trained skills into a single controller and adapt re-use the knowledge of those skills to solve new tasks. 
In line with previous work (\cite{extremeparkour}), we have shown that a policy resulting from standard distillation from multiple experts performs worse than the experts on their respective tasks and does not generalize to new scenarios. The performance loss can be explained by the fact that the experts choose different actions from the same (or similar) states, which makes the imitation problem ill-posed. The change of perception modality between experts and the student exacerbates the challenge since the distilled policy does not always have enough information to imitate the expert.

While the distilled policy does not offer sufficient robustness to be deployed in the real world, it does extract a nontrivial amount of knowledge from the experts, and its behavior needs only minor modifications to successfully solve all of the expert tasks. We show that this minor modification can be effectively achieved by fine-tuning the distilled policy with RL. Additionally, during the fine-tuning process, new tasks can be added for which no expert exists. 

It is tempting to directly train a policy from scratch on the final set of tasks, and in theory, nothing prevents this from being a suitable approach. In practice however, we find that this method is not practical. There are multiple issues related to the challenge of exploration. First, the policy needs to learn a diverse set of behaviors to solve the different tasks. We see that when the number of tasks increases, the RL process tends to focus on a subset of them and never recovers to find a suitable solution for the others. Second, even for the tasks where the policy finds a solution, it tends to lack specialization in each task. Instead, the policy learns a unique sub-optimal solution that can be applied across multiple tasks. Finally, extensive tuning of rewards, terrain, and curriculum is required to achieve a suitable solution even for the expert skills. When trained separately the effort is manageable since the tuning can be specialized to each task independently. However, when training all of the skills at the same time, a tuning decision for one task creates unintended consequences for others. Thus, achieving the desired quality of solutions for all tasks becomes untractable.

A suitable method for skill combination should allow training these skills separately and subsequently freeze them. The resulting controller should then retain the motion quality of those skills without the needed for further extensive tuning. Our approach does not entirely fulfill this requirement. While the distillation process uses frozen skills, the fine-tuning process changes the behavior of the policy and re-creates some of the challenges that appear when training from scratch. Since the skills were trained with slightly different rewards and termination, a new common setup is needed. Minor changes in the setup can lead to significant differences in the final behavior, favoring some of the terrains over others. The tuning effort is reduced compared to training from scratch but is nevertheless substantial.

In this work, we examine various methods to combine skills effectively. We attempt to reuse the hierarchical setup of our previous work but find that it does not scale well to the complex and unstructured terrains present in this work. Since the high-level policy can only select skills without any blending or interpolation, we find that the number of skills needs to be increased. Unfortunately, going from five to nine skills prevents the high-level policy from learning to use all of them. In our experiments, pre-training on expert terrains does not help with this problem.

Skill encoding, where motions of the different skills are encoded into a common latent space and a new policy is trained to control the robot through that latent space, has been studied extensively throughout this work. Our experiments show the advantages of such an approach in specific cases, but once again, scaling to an increasing number of skills and terrains proves to be a challenge. While this method allows us to efficiently train a policy for a single terrain requiring interpolation of two skills, once we add all of the terrains used during fine-tuning, the training efficiency deteriorates and eventually becomes inferior to our proposed approach.

\subsection{Perception modality}
Choosing the right perception modality is a considerable challenge when learning agile locomotion. In our previous work (\cite{hoeller2023anymal}), we developed a standalone perception module producing 3D and 2D representations of the terrain from depth images. Those representations were then used by the different policies. Following other works (\cite{extremeparkour, zhuang2023robotparkourlearning}), we choose to directly use depth images as input to the locomotion policy. This end-to-end approach has both advantages and disadvantages compared to the modularized approach. 

On one hand, having specialized modules requires the definition of interfaces, which can lead to a loss of information and capability. For example, the perception module of (\cite{hoeller2023anymal}) is predicting a 3D point cloud, which then needs to be converted to an elevation map. This conversion significantly deteriorates the information contained in the map. Additionally, every module requires an extensive development effort and expertise, while the end-to-end approach is more straightforward to implement. 

On the other hand, the end-to-end approach is harder to control and understand. Untangling different parts of the pipeline and attributing their errors to the final behavior of the robot becomes untractable. When the robot missteps or hits an obstacle, it is not possible to differentiate perception issues or sub-optimal locomotion behaviors. Furthermore, the tuning effort is significantly increased since it is not possible to fine-tune components separately.

\subsection{Possible policy improvements}
On top of the issues related to training and tuning described above, some limitations remain in the behavior learned by the policy. First, some of the robot's motions lack precision. On sparse terrains such as \textit{Beams} or \textit{Stepping-stones}, the policy tends to misstep and quickly recover rather than find a proper foothold directly. During deployment on unstructured terrains, the robot's feet often stumble on obstacles before overcoming them. Similarly, the robot sometimes needs to hit a table or another overhanging obstacle before initiating the crouching motion. These behaviors lead us to conclude that the policy learns to understand the presence of obstacles using proprioceptive information when perception fails. This is a great emerging behavior that significantly increases robustness, but by improving perception, we could reduce the need to use proprioception and improve the quality of the resulting motion. In particular, increasing the resolution and improving the sim-to-real transfer of the depth cameras while reducing the effects of the noise model can lead to significant improvements in the behavior.

Another drawback of the current policy is its limited use of memory. While in Sec. \ref{subsec:experiments/effect-of-nn-architecture}, we have shown that having a memory mechanism is crucial to learning the required behaviors, we see a lack of longer-term memory. As an example, we let the robot slowly approach a box until it is within the near-field clipping range of the camera. If the robot is commanded to move forward immediately, it will climb on the box directly, while if we let it stand for a few seconds, it will hit the box. This shows a limitation of the LSTM architecture with a hidden state that gets updated at every time-step. The relevant information about the box gets diluted while the robot is standing, even though no new information is coming in. A Transformer architecture with an attention mechanism (\cite{attentionisallyouneed}) could be beneficial for such scenarios.
 
% •	Success: high performance on all terrains, proper use of depth images where memory is required, good sim-to-real transfer, surprising robustness in real-life

% •	Fails: high effort is required to design all training terrains, once RL is added tuning is needed again, some terrains are still out of distribution (probably, not sure which ones), behavior is not optimal, and could be more gentle/precise
\section{Conclusion}
This work focuses on developing a general locomotion policy combining the agility of different locomotion skills into a single controller capable of tackling a large variety of obstacles in unstructured terrains. To that end, we develop a three-staged pipeline, where first individual skills are trained separately. Then the different skills are distilled into a single policy. Finally, the distilled policy is fine-tuned using RL on a new set of terrains, including the ones used to train the skills, but also 3D scans of real-world search-and-rescue training grounds. We use depth images as the only perceptive input to the policy. This reduces the dependency on state estimation and mapping, which tend to be unreliable in such scenarios. During training, we develop a custom depth noise model, which allows the transfer of the policy to the real world. We show that with our policy, the robot is capable of reaching places that were previously considered outside the capabilities of a legged robot. Furthermore, the capabilities of the policy can be continuously extended by adding new terrains to the training set and using repeated fine-tuning.

We compare different methods previously used to combine skills and find that our approach outperforms others when increasing the number of skills and terrains.

Nevertheless, both the qualitative and quantitative performance of the policy can be improved. In particular future work should revisit the noise model, making it, if possible, less aggressive to provide more precise information to the policy. Furthermore, the long-term memory of the policy can be improved by switching to another neural network architecture. Finally, the performance of the policy after distillation can be improved using a method capable of learning from a contradicting set of experts.

% \vspace{-2mm}
\subsubsection*{Funding statement}:
The project was funded by NVIDIA, the Swiss National Science Foundation (SNF) through the National Centre of Competence in Research Robotics, the European Research Council (ERC) under the European Union’s Horizon 2020 research and innovation program grant agreement No 852044 and No 780883. The work has been conducted as part of ANYmal Research, a community to advance legged robotics.

\clearpage 
\bibliographystyle{SageH}
\bibliography{references}

\begin{thebibliography}{38}
\providecommand{\natexlab}[1]{#1}
\providecommand{\url}[1]{\texttt{#1}}
\providecommand{\urlprefix}{URL }
\expandafter\ifx\csname urlstyle\endcsname\relax
  \providecommand{\doi}[1]{DOI:\discretionary{}{}{}#1}\else
  \providecommand{\doi}{DOI:\discretionary{}{}{}\begingroup \urlstyle{rm}\Url}\fi

\bibitem[{Agarwal et~al.(2022)Agarwal, Kumar, Malik and Pathak}]{agarwal2022legged}
Agarwal A, Kumar A, Malik J and Pathak D (2022) Legged locomotion in challenging terrains using egocentric vision.
\newblock In: \emph{6th Annual Conference on Robot Learning}.
\newblock \urlprefix\url{https://openreview.net/forum?id=Re3NjSwf0WF}.

\bibitem[{Alayrac et~al.(2022)Alayrac, Donahue, Luc, Miech, Barr, Hasson, Lenc, Mensch, Millican, Reynolds, Ring, Rutherford, Cabi, Han, Gong, Samangooei, Monteiro, Menick, Borgeaud, Brock, Nematzadeh, Sharifzadeh, Binkowski, Barreira, Vinyals, Zisserman and Simonyan}]{alayrac2022flamingovisuallanguagemodel}
Alayrac JB, Donahue J, Luc P, Miech A, Barr I, Hasson Y, Lenc K, Mensch A, Millican K, Reynolds M, Ring R, Rutherford E, Cabi S, Han T, Gong Z, Samangooei S, Monteiro M, Menick J, Borgeaud S, Brock A, Nematzadeh A, Sharifzadeh S, Binkowski M, Barreira R, Vinyals O, Zisserman A and Simonyan K (2022) Flamingo: a visual language model for few-shot learning.
\newblock \urlprefix\url{https://arxiv.org/abs/2204.14198}.

\bibitem[{Bohez et~al.(2022)Bohez, Tunyasuvunakool, Brakel, Sadeghi, Hasenclever, Tassa, Parisotto, Humplik, Haarnoja, Hafner, Wulfmeier, Neunert, Moran, Siegel, Huber, Romano, Batchelor, Casarini, Merel, Hadsell and Heess}]{bohez2022imitate}
Bohez S, Tunyasuvunakool S, Brakel P, Sadeghi F, Hasenclever L, Tassa Y, Parisotto E, Humplik J, Haarnoja T, Hafner R, Wulfmeier M, Neunert M, Moran B, Siegel N, Huber A, Romano F, Batchelor N, Casarini F, Merel J, Hadsell R and Heess N (2022) Imitate and repurpose: Learning reusable robot movement skills from human and animal behaviors.
\newblock \urlprefix\url{https://arxiv.org/abs/2203.17138}.

\bibitem[{Brohan et~al.(2023)Brohan, Brown, Carbajal, Chebotar, Dabis, Finn, Gopalakrishnan, Hausman, Herzog, Hsu, Ibarz, Ichter, Irpan, Jackson, Jesmonth, Joshi, Julian, Kalashnikov, Kuang and Zitkovich}]{rt1}
Brohan A, Brown N, Carbajal J, Chebotar Y, Dabis J, Finn C, Gopalakrishnan K, Hausman K, Herzog A, Hsu J, Ibarz J, Ichter B, Irpan A, Jackson T, Jesmonth S, Joshi N, Julian R, Kalashnikov D, Kuang Y and Zitkovich B (2023) Rt-1: Robotics transformer for real-world control at scale.
\newblock \doi{10.15607/RSS.2023.XIX.025}.

\bibitem[{Brown et~al.(2020)Brown, Mann, Ryder, Subbiah, Kaplan, Dhariwal, Neelakantan, Shyam, Sastry, Askell, Agarwal, Herbert-Voss, Krueger, Henighan, Child, Ramesh, Ziegler, Wu, Winter, Hesse, Chen, Sigler, Litwin, Gray, Chess, Clark, Berner, McCandlish, Radford, Sutskever and Amodei}]{GPT}
Brown TB, Mann B, Ryder N, Subbiah M, Kaplan J, Dhariwal P, Neelakantan A, Shyam P, Sastry G, Askell A, Agarwal S, Herbert-Voss A, Krueger G, Henighan T, Child R, Ramesh A, Ziegler DM, Wu J, Winter C, Hesse C, Chen M, Sigler E, Litwin M, Gray S, Chess B, Clark J, Berner C, McCandlish S, Radford A, Sutskever I and Amodei D (2020) Language models are few-shot learners.
\newblock \urlprefix\url{https://arxiv.org/abs/2005.14165}.

\bibitem[{Caluwaerts et~al.(2023)Caluwaerts, Iscen, Kew, Yu, Zhang, Freeman, Lee, Lee, Saliceti, Zhuang, Batchelor, Bohez, Casarini, Chen, Cortes, Coumans, Dostmohamed, Dulac-Arnold, Escontrela, Frey, Hafner, Jain, Jyenis, Kuang, Lee, Luu, Nachum, Oslund, Powell, Reyes, Romano, Sadeghi, Sloat, Tabanpour, Zheng, Neunert, Hadsell, Heess, Nori, Seto, Parada, Sindhwani, Vanhoucke and Tan}]{caluwaerts2023barkour}
Caluwaerts K, Iscen A, Kew JC, Yu W, Zhang T, Freeman D, Lee KH, Lee L, Saliceti S, Zhuang V, Batchelor N, Bohez S, Casarini F, Chen JE, Cortes O, Coumans E, Dostmohamed A, Dulac-Arnold G, Escontrela A, Frey E, Hafner R, Jain D, Jyenis B, Kuang Y, Lee E, Luu L, Nachum O, Oslund K, Powell J, Reyes D, Romano F, Sadeghi F, Sloat R, Tabanpour B, Zheng D, Neunert M, Hadsell R, Heess N, Nori F, Seto J, Parada C, Sindhwani V, Vanhoucke V and Tan J (2023) Barkour: Benchmarking animal-level agility with quadruped robots.

\bibitem[{Cheng et~al.(2024)Cheng, Shi, Agarwal and Pathak}]{extremeparkour}
Cheng X, Shi K, Agarwal A and Pathak D (2024) Extreme parkour with legged robots.
\newblock In: \emph{2024 IEEE International Conference on Robotics and Automation (ICRA)}. pp. 11443--11450.
\newblock \doi{10.1109/ICRA57147.2024.10610200}.

\bibitem[{Choi et~al.(2023)Choi, Ji, Park, Kim, Mun, Lee and Hwangbo}]{choi_23}
Choi S, Ji G, Park J, Kim H, Mun J, Lee JH and Hwangbo J (2023) Learning quadrupedal locomotion on deformable terrain.
\newblock \emph{Science Robotics} 8(74): eade2256.
\newblock \doi{10.1126/scirobotics.ade2256}.
\newblock \urlprefix\url{https://www.science.org/doi/abs/10.1126/scirobotics.ade2256}.

\bibitem[{Collaboration et~al.(2024)Collaboration, O'Neill, Rehman, Gupta, Maddukuri, Gupta, Padalkar, Lee, Pooley, Gupta, Mandlekar, Jain, Tung, Bewley, Herzog, Irpan, Khazatsky, Rai, Gupta, Wang, Kolobov, Singh, Garg, Kembhavi, Xie, Brohan, Raffin, Sharma, Yavary, Jain, Balakrishna, Wahid, Burgess-Limerick, Kim, Schölkopf, Wulfe, Ichter, Lu, Xu, Le, Finn, Wang, Xu, Chi, Huang, Chan, Agia, Pan, Fu, Devin, Xu, Morton, Driess, Chen, Pathak, Shah, Büchler, Jayaraman, Kalashnikov, Sadigh, Johns, Foster, Liu, Ceola, Xia, Zhao, Frujeri, Stulp, Zhou, Sukhatme, Salhotra, Yan, Feng, Schiavi, Berseth, Kahn, Yang, Wang, Su, Fang, Shi, Bao, Amor, Christensen, Furuta, Bharadhwaj, Walke, Fang, Ha, Mordatch, Radosavovic, Leal, Liang, Abou-Chakra, Kim, Drake, Peters, Schneider, Hsu, Vakil, Bohg, Bingham, Wu, Gao, Hu, Wu, Wu, Sun, Luo, Gu, Tan, Oh, Wu, Lu, Yang, Malik, Silvério, Hejna, Booher, Tompson, Yang, Salvador, Lim, Han, Wang, Rao, Pertsch, Hausman, Go, Gopalakrishnan, Goldberg, Byrne, Oslund, Kawaharazuka, Black,
  Lin, Zhang, Ehsani, Lekkala, Ellis, Rana, Srinivasan, Fang, Singh, Zeng, Hatch, Hsu, Itti, Chen, Pinto, Fei-Fei, Tan, Fan, Ott, Lee, Weihs, Chen, Lepert, Memmel, Tomizuka, Itkina, Castro, Spero, Du, Ahn, Yip, Zhang, Ding, Heo, Srirama, Sharma, Kim, Kanazawa, Hansen, Heess, Joshi, Suenderhauf, Liu, Palo, Shafiullah, Mees, Kroemer, Bastani, Sanketi, Miller, Yin, Wohlhart, Xu, Fagan, Mitrano, Sermanet, Abbeel, Sundaresan, Chen, Vuong, Rafailov, Tian, Doshi, Mart'in-Mart'in, Baijal, Scalise, Hendrix, Lin, Qian, Zhang, Mendonca, Shah, Hoque, Julian, Bustamante, Kirmani, Levine, Lin, Moore, Bahl, Dass, Sonawani, Tulsiani, Song, Xu, Haldar, Karamcheti, Adebola, Guist, Nasiriany, Schaal, Welker, Tian, Ramamoorthy, Dasari, Belkhale, Park, Nair, Mirchandani, Osa, Gupta, Harada, Matsushima, Xiao, Kollar, Yu, Ding, Davchev, Zhao, Armstrong, Darrell, Chung, Jain, Kumar, Vanhoucke, Zhan, Zhou, Burgard, Chen, Chen, Wang, Zhu, Geng, Liu, Liangwei, Li, Pang, Lu, Ma, Kim, Chebotar, Zhou, Zhu, Wu, Xu, Wang, Bisk, Dou, Cho,
  Lee, Cui, Cao, Wu, Tang, Zhu, Zhang, Jiang, Li, Li, Iwasawa, Matsuo, Ma, Xu, Cui, Zhang, Fu and Lin}]{rtx}
Collaboration E, O'Neill A, Rehman A, Gupta A, Maddukuri A, Gupta A, Padalkar A, Lee A, Pooley A, Gupta A, Mandlekar A, Jain A, Tung A, Bewley A, Herzog A, Irpan A, Khazatsky A, Rai A, Gupta A, Wang A, Kolobov A, Singh A, Garg A, Kembhavi A, Xie A, Brohan A, Raffin A, Sharma A, Yavary A, Jain A, Balakrishna A, Wahid A, Burgess-Limerick B, Kim B, Schölkopf B, Wulfe B, Ichter B, Lu C, Xu C, Le C, Finn C, Wang C, Xu C, Chi C, Huang C, Chan C, Agia C, Pan C, Fu C, Devin C, Xu D, Morton D, Driess D, Chen D, Pathak D, Shah D, Büchler D, Jayaraman D, Kalashnikov D, Sadigh D, Johns E, Foster E, Liu F, Ceola F, Xia F, Zhao F, Frujeri FV, Stulp F, Zhou G, Sukhatme GS, Salhotra G, Yan G, Feng G, Schiavi G, Berseth G, Kahn G, Yang G, Wang G, Su H, Fang HS, Shi H, Bao H, Amor HB, Christensen HI, Furuta H, Bharadhwaj H, Walke H, Fang H, Ha H, Mordatch I, Radosavovic I, Leal I, Liang J, Abou-Chakra J, Kim J, Drake J, Peters J, Schneider J, Hsu J, Vakil J, Bohg J, Bingham J, Wu J, Gao J, Hu J, Wu J, Wu J, Sun J, Luo J, Gu
  J, Tan J, Oh J, Wu J, Lu J, Yang J, Malik J, Silvério J, Hejna J, Booher J, Tompson J, Yang J, Salvador J, Lim JJ, Han J, Wang K, Rao K, Pertsch K, Hausman K, Go K, Gopalakrishnan K, Goldberg K, Byrne K, Oslund K, Kawaharazuka K, Black K, Lin K, Zhang K, Ehsani K, Lekkala K, Ellis K, Rana K, Srinivasan K, Fang K, Singh KP, Zeng KH, Hatch K, Hsu K, Itti L, Chen LY, Pinto L, Fei-Fei L, Tan L, Fan LJ, Ott L, Lee L, Weihs L, Chen M, Lepert M, Memmel M, Tomizuka M, Itkina M, Castro MG, Spero M, Du M, Ahn M, Yip MC, Zhang M, Ding M, Heo M, Srirama MK, Sharma M, Kim MJ, Kanazawa N, Hansen N, Heess N, Joshi NJ, Suenderhauf N, Liu N, Palo ND, Shafiullah NMM, Mees O, Kroemer O, Bastani O, Sanketi PR, Miller PT, Yin P, Wohlhart P, Xu P, Fagan PD, Mitrano P, Sermanet P, Abbeel P, Sundaresan P, Chen Q, Vuong Q, Rafailov R, Tian R, Doshi R, Mart'in-Mart'in R, Baijal R, Scalise R, Hendrix R, Lin R, Qian R, Zhang R, Mendonca R, Shah R, Hoque R, Julian R, Bustamante S, Kirmani S, Levine S, Lin S, Moore S, Bahl S, Dass S,
  Sonawani S, Tulsiani S, Song S, Xu S, Haldar S, Karamcheti S, Adebola S, Guist S, Nasiriany S, Schaal S, Welker S, Tian S, Ramamoorthy S, Dasari S, Belkhale S, Park S, Nair S, Mirchandani S, Osa T, Gupta T, Harada T, Matsushima T, Xiao T, Kollar T, Yu T, Ding T, Davchev T, Zhao TZ, Armstrong T, Darrell T, Chung T, Jain V, Kumar V, Vanhoucke V, Zhan W, Zhou W, Burgard W, Chen X, Chen X, Wang X, Zhu X, Geng X, Liu X, Liangwei X, Li X, Pang Y, Lu Y, Ma YJ, Kim Y, Chebotar Y, Zhou Y, Zhu Y, Wu Y, Xu Y, Wang Y, Bisk Y, Dou Y, Cho Y, Lee Y, Cui Y, Cao Y, Wu YH, Tang Y, Zhu Y, Zhang Y, Jiang Y, Li Y, Li Y, Iwasawa Y, Matsuo Y, Ma Z, Xu Z, Cui ZJ, Zhang Z, Fu Z and Lin Z (2024) Open x-embodiment: Robotic learning datasets and rt-x models.
\newblock \urlprefix\url{https://arxiv.org/abs/2310.08864}.

\bibitem[{Driess et~al.(2023)Driess, Xia, Sajjadi, Lynch, Chowdhery, Ichter, Wahid, Tompson, Vuong, Yu, Huang, Chebotar, Sermanet, Duckworth, Levine, Vanhoucke, Hausman, Toussaint, Greff, Zeng, Mordatch and Florence}]{driess2023palme}
Driess D, Xia F, Sajjadi MSM, Lynch C, Chowdhery A, Ichter B, Wahid A, Tompson J, Vuong Q, Yu T, Huang W, Chebotar Y, Sermanet P, Duckworth D, Levine S, Vanhoucke V, Hausman K, Toussaint M, Greff K, Zeng A, Mordatch I and Florence P (2023) Palm-e: An embodied multimodal language model.
\newblock In: \emph{arXiv preprint arXiv:2303.03378}.

\bibitem[{Fankhauser et~al.(2014)Fankhauser, Bloesch, Gehring, Hutter and Siegwart}]{frankhauserElevationMap}
Fankhauser P, Bloesch M, Gehring C, Hutter M and Siegwart R (2014) Robot-centric elevation mapping with uncertainty estimates.
\newblock In: \emph{Mobile Service Robotics}. pp. 433--440.
\newblock \doi{10.1142/9789814623353_0051}.
\newblock \urlprefix\url{https://www.worldscientific.com/doi/abs/10.1142/9789814623353_0051}.

\bibitem[{Gangapurwala et~al.(2022)Gangapurwala, Geisert, Orsolino, Fallon and Havoutis}]{rloc}
Gangapurwala S, Geisert M, Orsolino R, Fallon M and Havoutis I (2022) Rloc: Terrain-aware legged locomotion using reinforcement learning and optimal control.
\newblock \emph{IEEE Transactions on Robotics} 38(5): 2908–2927.
\newblock \doi{10.1109/tro.2022.3172469}.
\newblock \urlprefix\url{http://dx.doi.org/10.1109/TRO.2022.3172469}.

\bibitem[{Ge et~al.(2023)Ge, Macaluso, Li, Luo and Wang}]{PAFF}
Ge Y, Macaluso A, Li L, Luo P and Wang X (2023) Policy adaptation from foundation model feedback.
\newblock pp. 19059--19069.
\newblock \doi{10.1109/CVPR52729.2023.01827}.

\bibitem[{Grandia et~al.(2022)Grandia, Jenelten, Yang, Farshidian and Hutter}]{grandia2022perceptivelocomotionnonlinearmodel}
Grandia R, Jenelten F, Yang S, Farshidian F and Hutter M (2022) Perceptive locomotion through nonlinear model predictive control.
\newblock \urlprefix\url{https://arxiv.org/abs/2208.08373}.

\bibitem[{Gupta et~al.(2022)Gupta, Fan, Ganguli and Fei-Fei}]{gupta2022metamorph}
Gupta A, Fan L, Ganguli S and Fei-Fei L (2022) Metamorph: Learning universal controllers with transformers.

\bibitem[{Han et~al.(2024)Han, Zhu, Sheng, Zhang, Li, Zhang, Zhang, Liu, Zhou, Zhao, Li, Zhang, Wang, Chi, Li, Zhu, Xiang, Teng and Zhang}]{tencent}
Han L, Zhu Q, Sheng J, Zhang C, Li T, Zhang Y, Zhang H, Liu Y, Zhou C, Zhao R, Li J, Zhang Y, Wang R, Chi W, Li X, Zhu Y, Xiang L, Teng X and Zhang Z (2024) Lifelike agility and play in quadrupedal robots using reinforcement learning and generative pre-trained models.
\newblock \emph{Nature Machine Intelligence} 6(787–798).
\newblock \doi{10.1038/s42256-024-00861-3}.
\newblock \urlprefix\url{https://www.nature.com/articles/s42256-024-00861-3}.

\bibitem[{Hochreiter and Schmidhuber(1997)}]{LSTM}
Hochreiter S and Schmidhuber J (1997) {Long Short-Term Memory}.
\newblock \emph{Neural Computation} 9(8): 1735--1780.
\newblock \doi{10.1162/neco.1997.9.8.1735}.
\newblock \urlprefix\url{https://doi.org/10.1162/neco.1997.9.8.1735}.

\bibitem[{Hoeller et~al.(2023)Hoeller, Rudin, Sako and Hutter}]{hoeller2023anymal}
Hoeller D, Rudin N, Sako D and Hutter M (2023) Anymal parkour: Learning agile navigation for quadrupedal robots.

\bibitem[{Jenelten et~al.(2022)Jenelten, Grandia, Farshidian and Hutter}]{jenelten2022Tamols}
Jenelten F, Grandia R, Farshidian F and Hutter M (2022) Tamols: Terrain-aware motion optimization for legged systems.
\newblock \emph{IEEE Transactions on Robotics} 38(6): 3395--3413.

\bibitem[{Jenelten et~al.(2024)Jenelten, He, Farshidian and Hutter}]{deeptracking}
Jenelten F, He J, Farshidian F and Hutter M (2024) Dtc: Deep tracking control.
\newblock \emph{Science Robotics} 9(86).
\newblock \doi{10.1126/scirobotics.adh5401}.
\newblock \urlprefix\url{http://dx.doi.org/10.1126/scirobotics.adh5401}.

\bibitem[{Kang et~al.(2023)Kang, Cheng, Zamora, Zargarbashi and Coros}]{kang_RL+MPC}
Kang D, Cheng J, Zamora M, Zargarbashi F and Coros S (2023) Rl + model-based control: Using on-demand optimal control to learn versatile legged locomotion.
\newblock \emph{IEEE Robotics and Automation Letters} 8(10): 6619--6626.
\newblock \doi{10.1109/LRA.2023.3307008}.

\bibitem[{Kim et~al.(2020)Kim, Carballo, Di~Carlo, Katz, Bledt, Lim and Kim}]{kim2020loco}
Kim D, Carballo D, Di~Carlo J, Katz B, Bledt G, Lim B and Kim S (2020) Vision aided dynamic exploration of unstructured terrain with a small-scale quadruped robot.
\newblock In: \emph{2020 IEEE International Conference on Robotics and Automation (ICRA)}. pp. 2464--2470.

\bibitem[{Lee et~al.(2020)Lee, Hwangbo, Wellhausen, Koltun and Hutter}]{Lee_2020}
Lee J, Hwangbo J, Wellhausen L, Koltun V and Hutter M (2020) Learning quadrupedal locomotion over challenging terrain.
\newblock \emph{Science Robotics} 5(47).
\newblock \doi{10.1126/scirobotics.abc5986}.
\newblock \urlprefix\url{http://dx.doi.org/10.1126/scirobotics.abc5986}.

\bibitem[{Luo et~al.(2024)Luo, Cao, Merel, Winkler, Huang, Kitani and Xu}]{luo2024universal}
Luo Z, Cao J, Merel J, Winkler A, Huang J, Kitani KM and Xu W (2024) Universal humanoid motion representations for physics-based control.
\newblock In: \emph{The Twelfth International Conference on Learning Representations}.
\newblock \urlprefix\url{https://openreview.net/forum?id=OrOd8PxOO2}.

\bibitem[{Miki et~al.(2022{\natexlab{a}})Miki, Lee, Hwangbo, Wellhausen, Koltun and Hutter}]{Miki_2022}
Miki T, Lee J, Hwangbo J, Wellhausen L, Koltun V and Hutter M (2022{\natexlab{a}}) Learning robust perceptive locomotion for quadrupedal robots in the wild.
\newblock \emph{Science Robotics} 7(62).
\newblock \doi{10.1126/scirobotics.abk2822}.
\newblock \urlprefix\url{http://dx.doi.org/10.1126/scirobotics.abk2822}.

\bibitem[{Miki et~al.(2022{\natexlab{b}})Miki, Wellhausen, Grandia, Jenelten, Homberger and Hutter}]{miki2022elevationmappinglocomotionnavigation}
Miki T, Wellhausen L, Grandia R, Jenelten F, Homberger T and Hutter M (2022{\natexlab{b}}) Elevation mapping for locomotion and navigation using gpu.
\newblock \urlprefix\url{https://arxiv.org/abs/2204.12876}.

\bibitem[{Mittal et~al.(2024)Mittal, Rudin, Klemm, Allshire and Hutter}]{mittal2024symmetryconsiderationslearningtask}
Mittal M, Rudin N, Klemm V, Allshire A and Hutter M (2024) Symmetry considerations for learning task symmetric robot policies.
\newblock \urlprefix\url{https://arxiv.org/abs/2403.04359}.

\bibitem[{Perlin(2002)}]{PerlinNoise}
Perlin K (2002) Improving noise.
\newblock In: \emph{Proceedings of the 29th Annual Conference on Computer Graphics and Interactive Techniques}, SIGGRAPH '02. New York, NY, USA: Association for Computing Machinery.
\newblock ISBN 1581135211, p. 681–682.
\newblock \doi{10.1145/566570.566636}.
\newblock \urlprefix\url{https://doi.org/10.1145/566570.566636}.

\bibitem[{Ross et~al.(2011)Ross, Gordon and Bagnell}]{dagger}
Ross S, Gordon GJ and Bagnell JA (2011) A reduction of imitation learning and structured prediction to no-regret online learning.

\bibitem[{Rudin et~al.(2022{\natexlab{a}})Rudin, Hoeller, Bjelonic and Hutter}]{rudin2022advanced}
Rudin N, Hoeller D, Bjelonic M and Hutter M (2022{\natexlab{a}}) Advanced skills by learning locomotion and local navigation end-to-end.

\bibitem[{Rudin et~al.(2022{\natexlab{b}})Rudin, Hoeller, Reist and Hutter}]{rudin2022learning}
Rudin N, Hoeller D, Reist P and Hutter M (2022{\natexlab{b}}) Learning to walk in minutes using massively parallel deep reinforcement learning.

\bibitem[{Vaswani et~al.(2017)Vaswani, Shazeer, Parmar, Uszkoreit, Jones, Gomez, Kaiser and Polosukhin}]{attentionisallyouneed}
Vaswani A, Shazeer N, Parmar N, Uszkoreit J, Jones L, Gomez AN, Kaiser L and Polosukhin I (2017) Attention is all you need.
\newblock In: Guyon I, Luxburg UV, Bengio S, Wallach H, Fergus R, Vishwanathan S and Garnett R (eds.) \emph{Advances in Neural Information Processing Systems}, volume~30. Curran Associates, Inc.
\newblock \urlprefix\url{https://proceedings.neurips.cc/paper_files/paper/2017/file/3f5ee243547dee91fbd053c1c4a845aa-Paper.pdf}.

\bibitem[{Xie et~al.(2023)Xie, Da, Babich, Garg and de~Panne}]{glide}
Xie Z, Da X, Babich B, Garg A and de~Panne Mv (2023) Glide: Generalizable quadrupedal locomotion in diverse environments with a centroidal model.
\newblock In: LaValle SM, O'Kane JM, Otte M, Sadigh D and Tokekar P (eds.) \emph{Algorithmic Foundations of Robotics XV}. Cham: Springer International Publishing.
\newblock ISBN 978-3-031-21090-7, pp. 523--539.

\bibitem[{Yang et~al.(2022)Yang, Zhang, Hansen, Xu and Wang}]{yang2022learning}
Yang R, Zhang M, Hansen N, Xu H and Wang X (2022) Learning vision-guided quadrupedal locomotion end-to-end with cross-modal transformers.

\bibitem[{Yu et~al.(2022)Yu, Jain, Escontrela, Iscen, Xu, Coumans, Ha, Tan and Zhang}]{yu_visual_locomotion}
Yu W, Jain D, Escontrela A, Iscen A, Xu P, Coumans E, Ha S, Tan J and Zhang T (2022) Visual-locomotion: Learning to walk on complex terrains with vision.
\newblock In: Faust A, Hsu D and Neumann G (eds.) \emph{Proceedings of the 5th Conference on Robot Learning}, \emph{Proceedings of Machine Learning Research}, volume 164. PMLR, pp. 1291--1302.
\newblock \urlprefix\url{https://proceedings.mlr.press/v164/yu22a.html}.

\bibitem[{Zhang et~al.(2024)Zhang, Rudin, Hoeller and Hutter}]{zhang2024learningagilelocomotionrisky}
Zhang C, Rudin N, Hoeller D and Hutter M (2024) Learning agile locomotion on risky terrains.
\newblock \urlprefix\url{https://arxiv.org/abs/2311.10484}.

\bibitem[{Zhuang et~al.(2023)Zhuang, Fu, Wang, Atkeson, Schwertfeger, Finn and Zhao}]{zhuang2023robotparkourlearning}
Zhuang Z, Fu Z, Wang J, Atkeson C, Schwertfeger S, Finn C and Zhao H (2023) Robot parkour learning.
\newblock \urlprefix\url{https://arxiv.org/abs/2309.05665}.

\bibitem[{Zitkovich et~al.(2023)Zitkovich, Yu, Xu, Xu, Xiao, Xia, Wu, Wohlhart, Welker, Wahid, Vuong, Vanhoucke, Tran, Soricut, Singh, Singh, Sermanet, Sanketi, Salazar, Ryoo, Reymann, Rao, Pertsch, Mordatch, Michalewski, Lu, Levine, Lee, Lee, Leal, Kuang, Kalashnikov, Julian, Joshi, Irpan, brian ichter, Hsu, Herzog, Hausman, Gopalakrishnan, Fu, Florence, Finn, Dubey, Driess, Ding, Choromanski, Chen, Chebotar, Carbajal, Brown, Brohan, Arenas and Han}]{rt2}
Zitkovich B, Yu T, Xu S, Xu P, Xiao T, Xia F, Wu J, Wohlhart P, Welker S, Wahid A, Vuong Q, Vanhoucke V, Tran H, Soricut R, Singh A, Singh J, Sermanet P, Sanketi PR, Salazar G, Ryoo MS, Reymann K, Rao K, Pertsch K, Mordatch I, Michalewski H, Lu Y, Levine S, Lee L, Lee TWE, Leal I, Kuang Y, Kalashnikov D, Julian R, Joshi NJ, Irpan A, brian ichter, Hsu J, Herzog A, Hausman K, Gopalakrishnan K, Fu C, Florence P, Finn C, Dubey KA, Driess D, Ding T, Choromanski KM, Chen X, Chebotar Y, Carbajal J, Brown N, Brohan A, Arenas MG and Han K (2023) {RT}-2: Vision-language-action models transfer web knowledge to robotic control.
\newblock In: \emph{7th Annual Conference on Robot Learning}.
\newblock \urlprefix\url{https://openreview.net/forum?id=XMQgwiJ7KSX}.

\end{thebibliography}
\clearpage

\end{document}